\documentclass[10pt,twocolumn,letterpaper]{article}

\usepackage{times}
\usepackage{epsfig}
\usepackage{graphicx}
\usepackage{amsmath}
\usepackage{amssymb}
\usepackage{tikz}
\usepackage{standalone}
\usepackage{subfig}
\usepackage{multirow}
\usepackage[normalem]{ulem}

\usepackage{xcolor,colortbl}

\usepackage[pagebackref=true,breaklinks=true,letterpaper=true,colorlinks,bookmarks=false]{hyperref}

\usepackage{amsmath,amsfonts,bm}

\def\eqref#1{equation~\ref{#1}}
\def\1{\bm{1}}

\DeclareMathAlphabet{\mathsfit}{\encodingdefault}{\sfdefault}{m}{sl}
\SetMathAlphabet{\mathsfit}{bold}{\encodingdefault}{\sfdefault}{bx}{n}

\usepackage{url}
\usepackage[capitalise,nameinlink]{cleveref}

\renewcommand{\Cref}[1]{\cref{#1}}
\Crefname{equation}{Eq.}{Eqs.}
\Crefname{figure}{Fig.}{Figs.}
\Crefname{tabular}{Tab.}{Tabs.}
\Crefname{table}{Tab.}{Tabs.}
\Crefname{section}{Sec.}{Secs.}     % smart cross-referencing
\usepackage{paralist}
\usepackage{bbm}
\begin{document}

%%%%%%%%% TITLE
\title{Detecting Novelties with Empty Classes}
\author{Svenja Uhlemeyer\\
IZMD and Faculty of Mathematics and Natural Sciences\\
University of Wuppertal, Germany \\
{\tt\small uhlemeyer@math.uni-wuppertal.de} \\
\and
Julian Lienen \\
Department of Computer Science\\
Paderborn University, Germany \\
{\tt\small julian.lienen@upb.de} \\
\and
Eyke Hüllermeier \\
Institute for Informatics\\
LMU Munich, Germany \\
{\tt\small eyke@lmu.de} \\
\and
Hanno Gottschalk\\
Institute of Mathematics\\
Technical University Berlin, Germany \\
{\tt\small gottschalk@math.tu-berlin.de}
}

\maketitle
%\ificcvfinal\thispagestyle{empty}\fi

\newcommand{\fix}{\marginpar{FIX}}
\newcommand{\new}{\marginpar{NEW}}
\newcommand{\SU}[1]{\textcolor{black}{#1}}
\newcommand{\JL}[1]{\textcolor{green}{#1}}
\newcommand{\outjl}[1]{\textcolor{green}{  \sout{#1}}}

\maketitle

\begin{abstract}
    For open world applications, deep neural networks (DNNs) need to be aware of previously unseen data and adaptable to evolving environments. Furthermore, it is desirable to detect and learn novel classes which are not included in the DNNs underlying set of semantic classes in an unsupervised fashion. The method proposed in this article builds upon anomaly detection to retrieve out-of-distribution (OoD) data as candidates for new classes. We thereafter extend the DNN by $k$ empty classes and fine-tune it on the OoD data samples. To this end, we introduce two loss functions, which 1) entice the DNN to assign OoD samples to the empty classes and 2) to minimize the inner-class feature distances between them. Thus, instead of ground truth which contains labels for the different novel classes, the DNN obtains a single OoD label together with a distance matrix, which is computed in advance. We perform several experiments for image classification and semantic segmentation, which demonstrate that a DNN can extend its own semantic space by multiple classes without having access to ground truth.
\end{abstract}

\section{Introduction}

\begin{figure}[ht]
    \centering
    \captionsetup[subfloat]{labelformat=empty, position=top}
    \subfloat[image]{\includegraphics[width=0.235\textwidth]{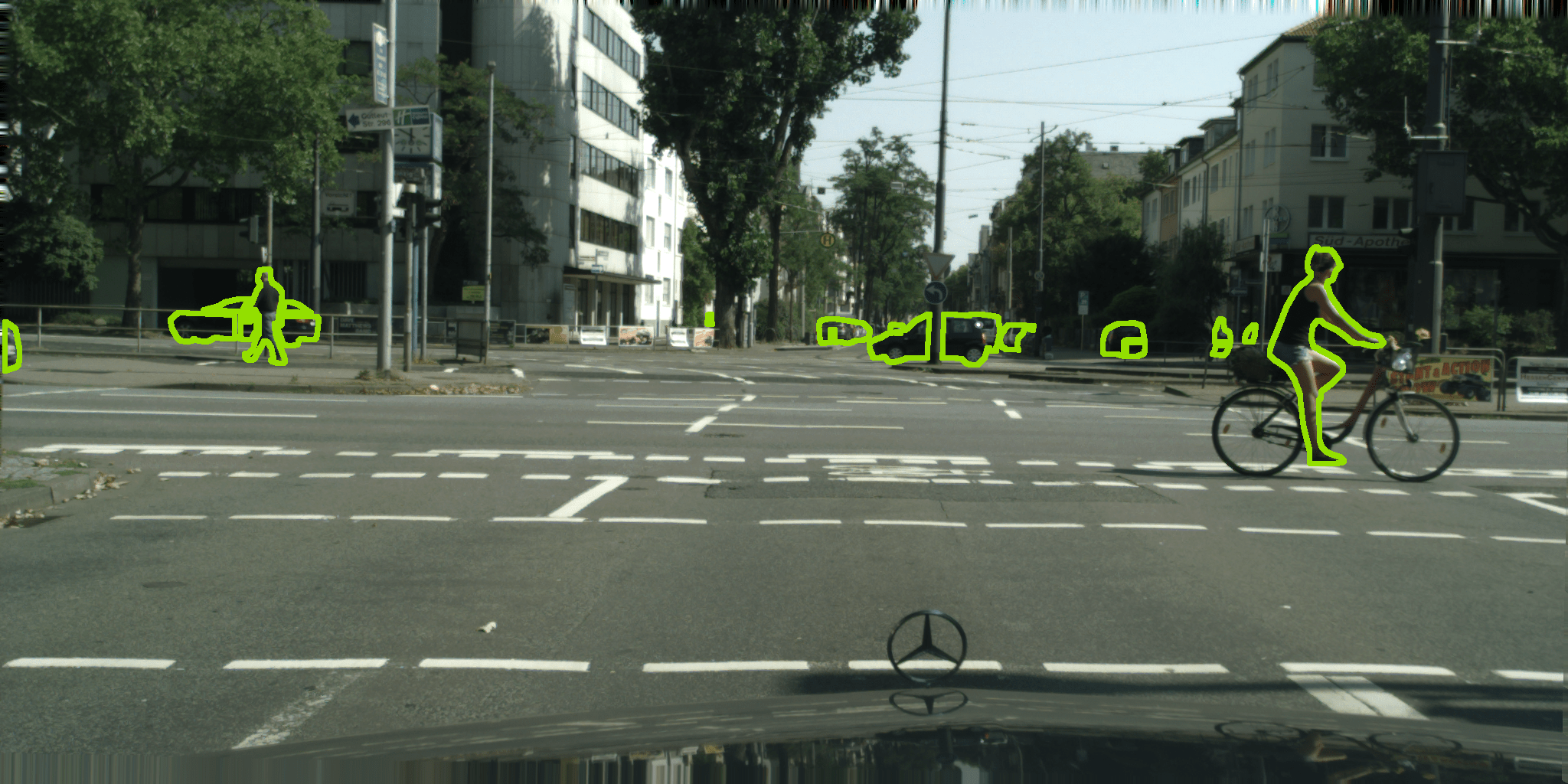}}\hfill
    \subfloat[ground truth]{\includegraphics[width=0.235\textwidth]{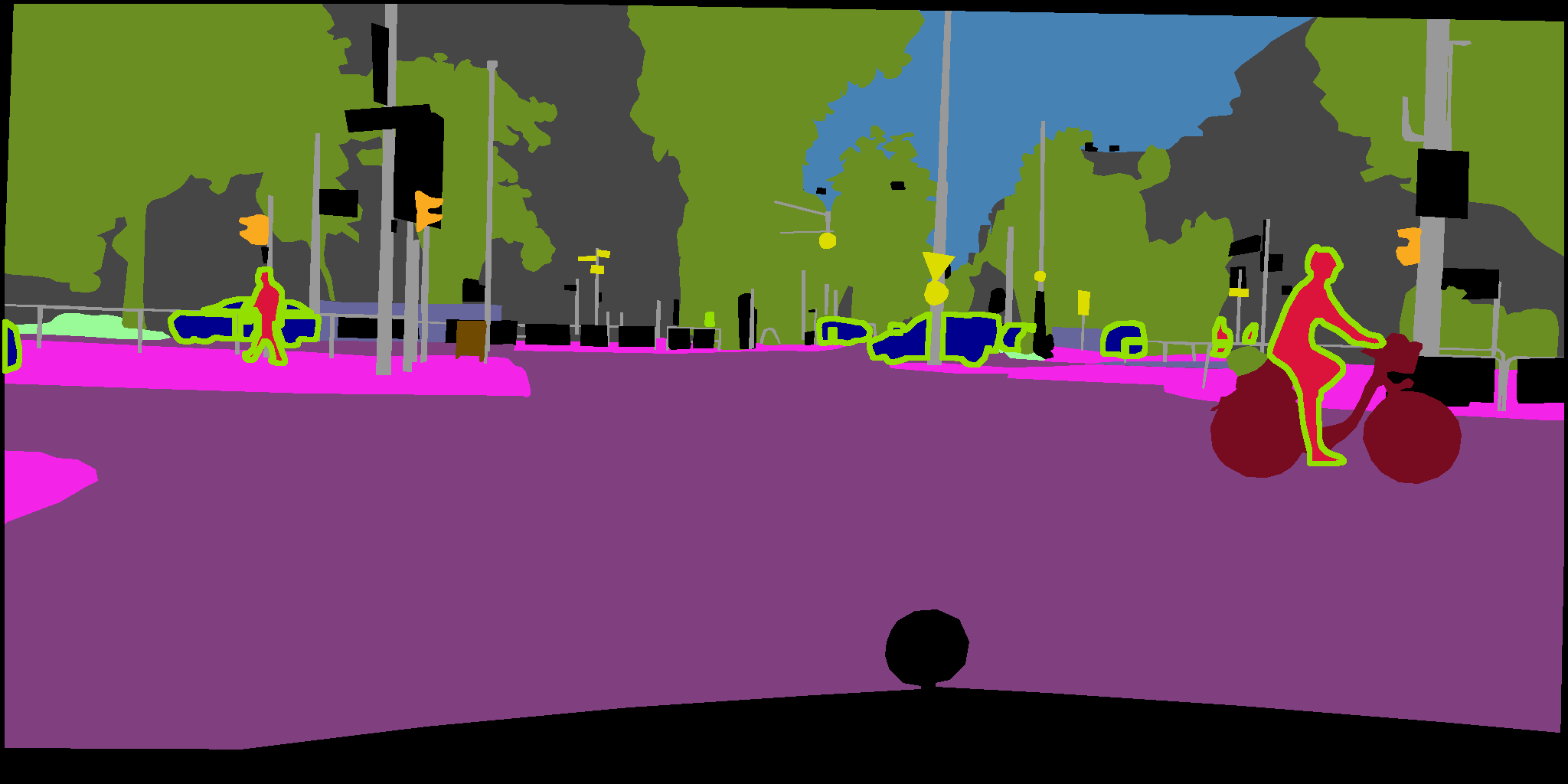}}\\[-1em]
    \captionsetup[subfloat]{labelformat=empty, position=bottom}
    \subfloat[baseline]{\includegraphics[width=0.235\textwidth]{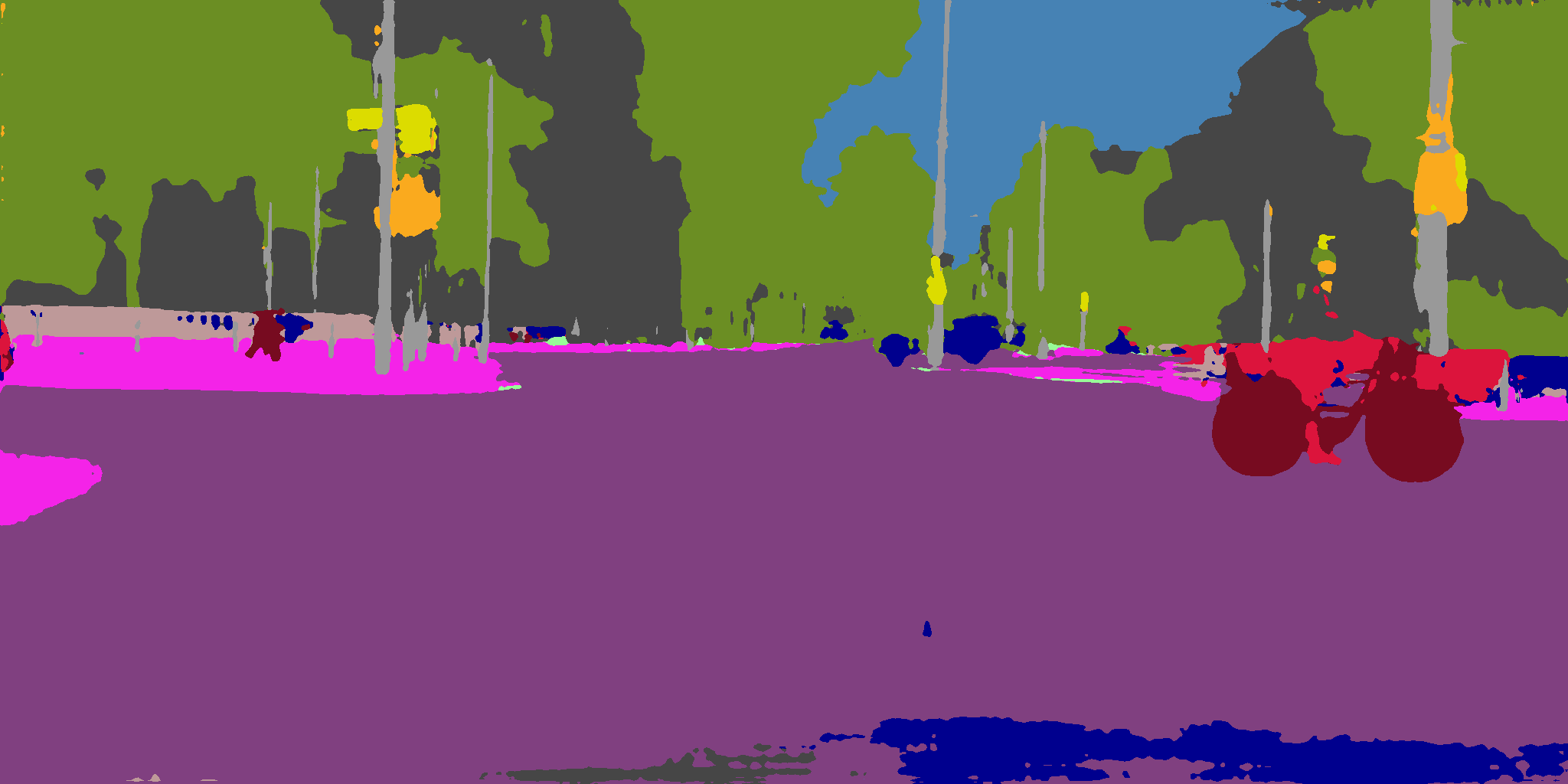}}\hfill
    \subfloat[ours]{\includegraphics[width=0.235\textwidth]{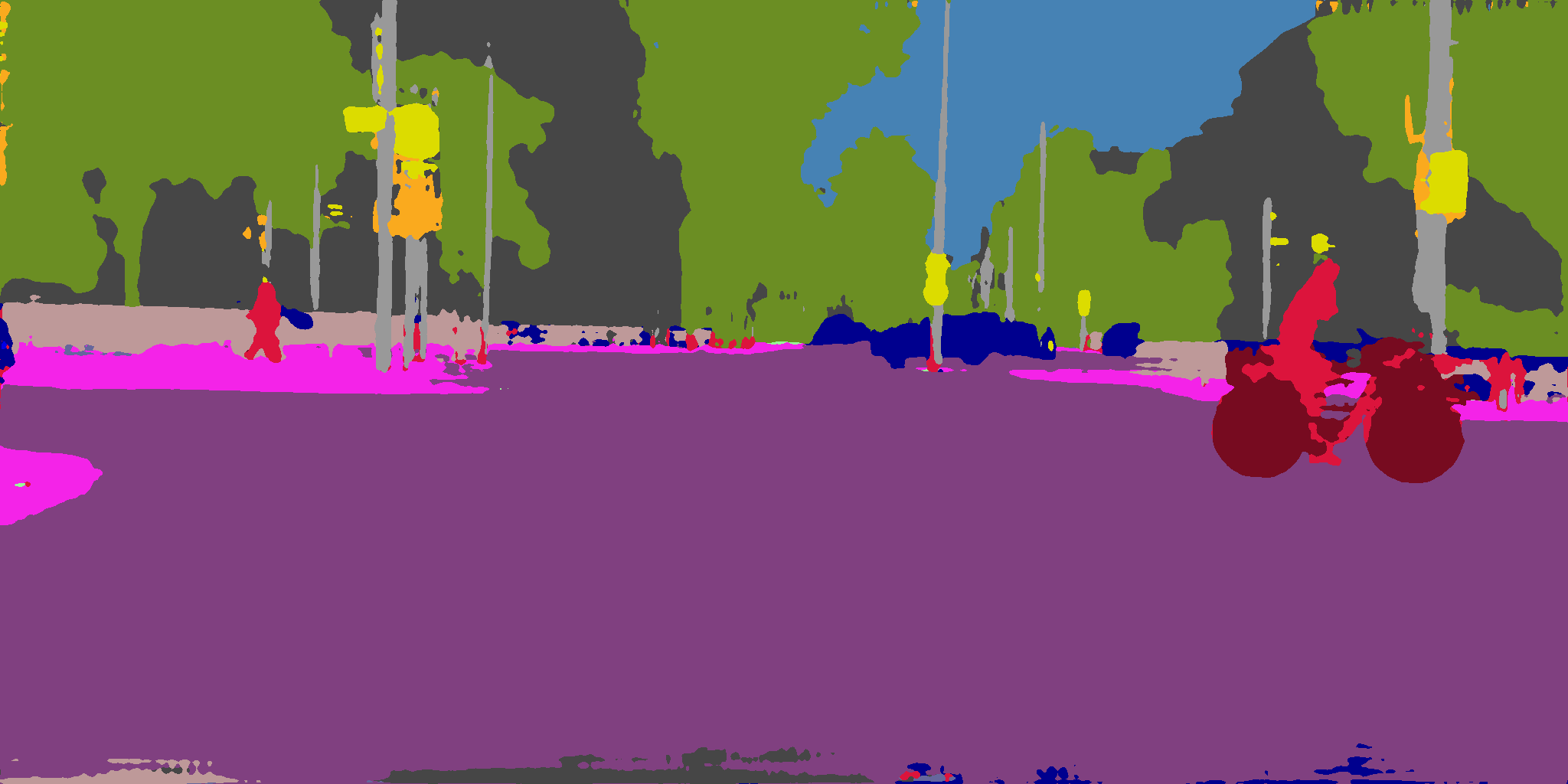}}
    \caption{Comparison of two segmentation DNNs which were extended by the classes \emph{human} and \emph{car}. While the segmentation masks are similar for the initial classes, the humans and cars are much better segmented by the DNN which was extended by our empty classes approach. The novel classes are marked with green contours in the image and ground truth.}
    \label{fig:page_1}
\end{figure}

For computer vision tasks such as image classification or semantic segmentation, deep neural networks (DNNs) learn to classify instances, either on a per image- or per pixel-level, into a limited number of predefined classes. State-of-the-art DNNs achieve high accuracy when trained in a supervised fashion and deployed in a closed world setting,
in which the learner is not confronted with out-of-distribution (OoD) data. In practice, however, concepts not seen at training time might occur, which is why so-called open world recognition \cite{Bendale2015TowardsOW} has emerged as a practically more relevant problem formulation.
It combines OoD detection with class-incremental learning, i.e.\ , retraining the model with newly observed classes. Nevertheless, methods of this kind are typically updated in a supervised fashion, commonly employing humans for annotation.
%Practically more relevant, open world recognition \cite{Bendale2015TowardsOW} combines OoD detection and class-incremental learning, however, mostly in a supervised context. This is, detected novel classes need to be annotated by humans to be learned incrementally. 

First attempts to learn in an unsupervised manner have been made to achieve cheaper labeling. In open world image classification, clustering methods like \emph{k-means} \cite{kmeans} or \emph{DBSCAN} \cite{dbscan} allow for an unsupervised labeling of instances in feature regions that appear to be novel. Approaches in this direction leverage such methods to obtain pseudo labels for detected OoD images \cite{He2021UnsupervisedCL,Shu2018UnseenCD}.

 \SU{However, the quality of these pseudo labels strongly depends on the clustering performance. Furthermore, the OoD candidates are assigned to fixed labels, which are likely to be noisy and thus unreliable, whereas in our method, they are put in relation to each other.} In open world semantic segmentation \cite{uhlemeyer2022towards,Nakajima2019IncrementalCD}, pseudo labeling on a per pixel-level is required, rendering the problem more complex. More recently, few-shot learning \cite{Cen_2021_ICCV}, where a model is trained to generalize well on novel classes with only few labeled examples, has also been proposed to deal with the lack of labeled data as another (semi-)supervised strategy.
%Clustering methods like \emph{K-means} \cite{kmeans} or \emph{DBSCAN} \cite{dbscan} serve as baseline approaches for open world image classification, since the numbering of the clusters is a sufficient labeling. 
%Approaches in this direction deploy clustering methods to obtain pseudo labels for detected OoD images \cite{He2021UnsupervisedCL,Shu2018UnseenCD}. For open world semantic segmentation \cite{uhlemeyer2022towards,Nakajima2019IncrementalCD}, pseudo labeling on pixel-level is required. Another (semi-)supervised strategy to handle the lack of labeled data is few-shot learning \cite{Cen_2021_ICCV}, where a model is trained to generalize well on novel classes with only few labeled examples.

\begin{figure*}[t]
    \centering
    \includegraphics[width=\textwidth]{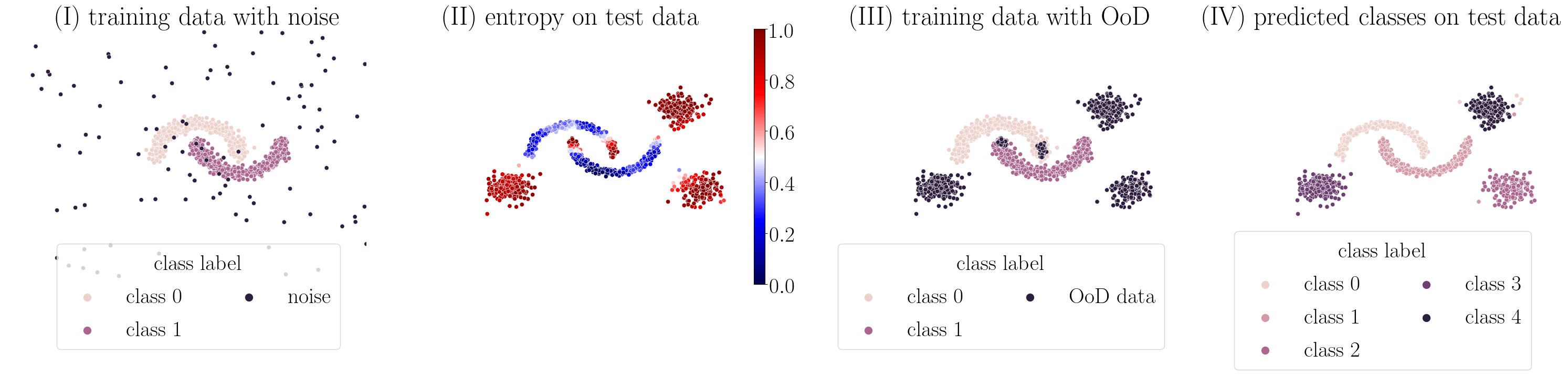}
    \caption{(I) A binary classification model is trained on two classes and additional noise data for entropy maximization. (II) OoD samples in the test data are obtained by entropy thresholding. (III) The training data is enriched with the OoD samples and a distance matrix, containing their pair-wise Euclidean distances. (IV) The model is class-incrementally extended by three novel classes.}
    \label{fig:toy_example}
\end{figure*}

% {\color{red} We introduce a new unsupervised approach for incrementally extending a DNN by novel classes, called \emph{dummy clustering}. Our method does not require any pseudo labels, as the clustering is not applied as a pre-processing step for the model extension, but it is directly integrated as additional loss function. We iteratively discover the number of \emph{dummy} classes, i.e.\ , auxiliary neurons in the DNN's output layer which constitute the novel classes. To this end, we track clustering evaluation metrics over a limited number of epochs (\emph{observation phase}) for an increasing number of dummies until some stopping criterion is fulfilled, then determine the quantity of novel classes via the elbow method \hg{HG: Zitat}.}

In our work, we introduce a new unsupervised approach for incrementally extending DNNs by capturing novel concepts in additional classes of hypothetical nature. To this end, we proceed from an initial model aware of hitherto known classes, which is augmented by an out-of-distribution detection mechanism to distinguish these classes from unknown categories. When treating additional data with potentially additional but unknown classes, we suggest to extend the model by additional auxiliary neurons in the DNN's output layer constituting the suspected novel classes to be recognized, which we dub \emph{empty classes}. To predict outcomes of these classes, our model is fine-tuned by a clustering loss that aims to recognize similar concepts for out-of-distribution data, allowing to flexibly adapt the learned feature representations to distinguish the already known classes from the new learning outcomes.

We conduct experiments on several datasets with increasing level of difficulty, starting with image classification of MNIST~\cite{LeCun1998GradientbasedLA} digits as well as the slightly more sophisticated data from FashionMNIST~\cite{Xiao2017FashionMNISTAN}. Next, we apply our approach to low- and medium-resolution images from the CIFAR10~\cite{Krizhevsky2009LearningML} and Animals10\footnote{\label{animals10}\href{https://www.kaggle.com/datasets/alessiocorrado99/animals10}{https://www.kaggle.com/datasets/alessiocorrado99/animals10}} dataset, respectively. Finally, we also adapt our method to the complex task of semantic segmentation of street scenes from the Cityscapes~\cite{Cordts2016TheCD} dataset. \SU{In three out of four image classification experiments, our method outperforms the baseline, where a DNN is fine-tuned on $k$-means labeled OoD data. Furthermore, our extended segmentation DNN achieves better results than the baseline~\cite{uhlemeyer2022towards} for the novel class \emph{car}, and significantly reduces the number of overlooked humans. See~\cref{fig:page_1} for an example.}

% \JL{Add that we achieve better performance.} %and the synthetic CARLA-WildLife~\cite{Maag_2022_ACCV} 

\section{Related Work}

%In the following, we revisit related literature on open world recognition for the problems of classification and particularly semantic segmentation, as well as highlight work in the field of unsupervised representation learning.

%\subsection{Open World Classification}

%\subsection{Open World Semantic Segmentation}

%\subsection{Unsupervised Representation Learning}

% Open World Recognition (open set + incremental learning)
Open world recognition~\cite{Bendale2015TowardsOW} refers to the problem of adapting a learning system to a non-delimitable and potentially constantly evolving target domain. As such, it combines the disciplines of open set learning~\cite{Scheirer2013TowardOS}, where incomplete knowledge over the target domain is assumed at training time, with incremental learning~\cite{Cauwenberghs2000IncrementalAD}, in which the model is updated by exploring additional target space regions at test time, thereby adapting to novel target information. Typically, open set recognition is formalized by specifying a novelty detector, a labeling process and an incremental learning function, allowing for a generalized characterization of such systems~\cite{Bendale2015TowardsOW}.

% Open World Classification
Most of previous approaches consider the open world recognition problem in the context of classification, where novel concepts are in form of previously unseen classes. While a plethora of methods has been proposed to tackle the individual sub-problems for classification problems, for which we refer to \cite{Parmar2021OpenworldML} for a more comprehensive overview, literature on holistic approaches for open world classification is rather scarce. In \cite{Shu2018UnseenCD}, a metric learning approach is used to distinguish between pairs of instances belonging to the same classes, allowing to detect instances that can not be mapped to known classes and being used to learn novel class concepts. Moreover, \cite{Nixon2020SemiSupervisedCD} suggests a semi-supervised learning approach that applies clustering on learned feature representations to reason about unknown classes. Related to this, \cite{Wang2020OpenWorldCD} describes a kernel method using an alternative loss formulation to learn embeddings to be clustered for class discovery. More recently, similar concepts have also been tailored to specific data modalities, such as tabular data \cite{Troisemaine2022AMF}.

% Open World Semantic Segmentation
In the domain of semantic segmentation, open world recognition is also covered under the term \emph{zero-shot semantic segmentation} \cite{Bucher2019ZeroShotSS}. To predict unseen categories for classified pixels, a wide range of methods leverage additional language-based context information \cite{Bucher2019ZeroShotSS, Xian2019SemanticPN, LIU2022OpenworldSS}. Besides enriching visual information by text, unsupervised methods, e.g.\ , employing clustering based on visual similarity \cite{uhlemeyer2022towards} or contrastive losses \cite{VanGansbeke2021UnsupervisedSS,Cen2021DeepML}, have also been considered. More recently, \cite{Cen2022OpenworldSS} adopts semantic segmentation based on LiDAR point clouds by augmenting conventional classifiers with predictors recognizing unknown classes, thereby enabling incremental learning.

% Unsupervised Representation Learning
In a more general context, unsupervised representation learning \cite{DBLP:journals/corr/RadfordMC15} constitutes a major challenge to generalize learning methods to unseen concepts. Methods of this kind are typically tailored to data modalities, e.g.\ , by specifying auxiliary tasks to be solved \cite{DBLP:conf/iclr/GidarisSK18,Yun2022PatchlevelRL}. In the domain of images, self-supervised learning approaches have emerged recently \cite{DBLP:conf/iccv/CaronTMJMBJ21,DBLP:conf/iclr/LiYZGXDYG22}, which commonly apply knowledge distillation between different networks, allowing for learning in a self-supervised fashion. Other methods including ideas stemming from metric \cite{DBLP:conf/nips/GrillSATRBDPGAP20} or contrastive learning \cite{DBLP:conf/nips/ChenKSNH20}.

\begin{figure*}[t]
    \centering
    \includegraphics[width=0.9\textwidth]{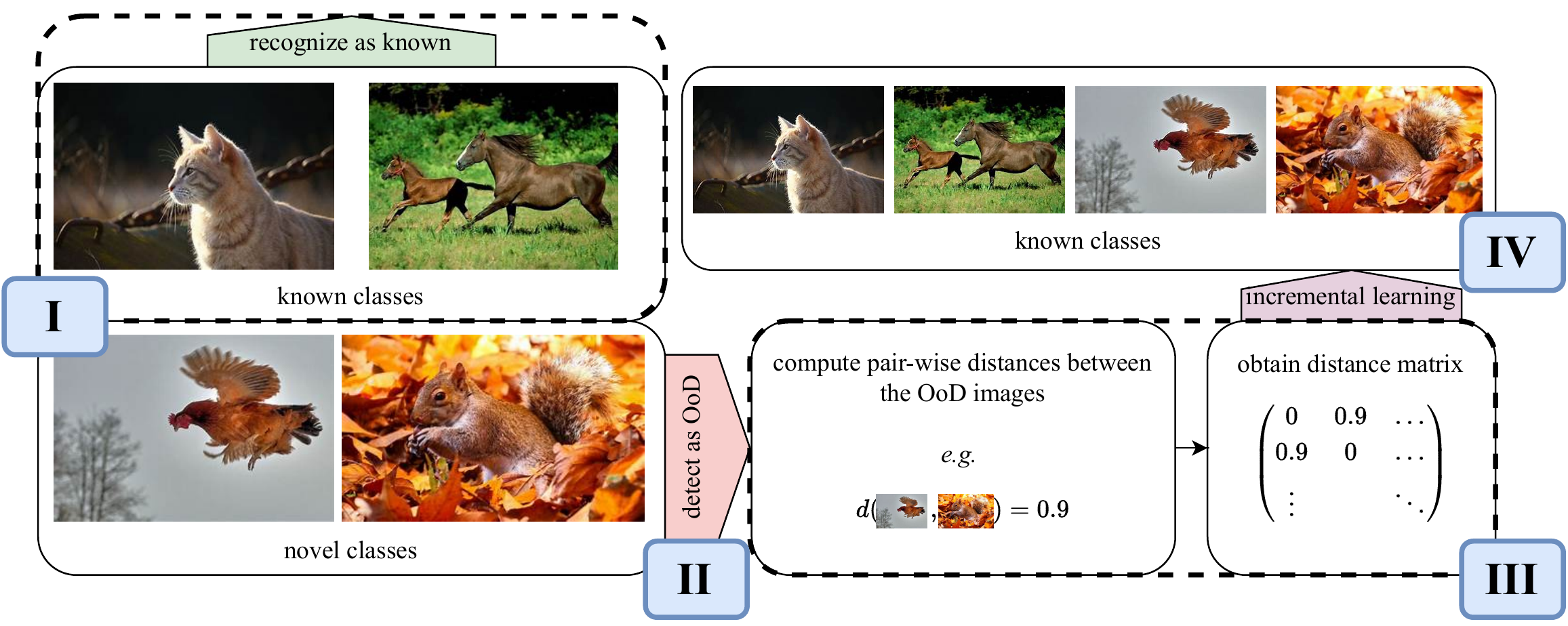}
    \caption{Open world recognition models must be able to recognize known classes while detecting OoD data from novel classes and furthermore to incrementally learn these novel classes. Instead of labeling the OoD samples, our method computes pair-wise distances between them, which serve as input for a clustering loss function.}\label{fig:method_overview}
\end{figure*}

% \section{\outhg{Method }Description \hg{of the Method}}
\section{Method Description}

In this section, we present our training framework for unsupervised class-incremental learning with empty classes. For the sake of brevity, all equations are introduced for image classification and adapted to semantic segmentation in~\cref{sec:semseg}. First, we give a motivating example in~\cref{fig:toy_example}, where we enrich data stemming from the TwoMoons dataset\footnote{\label{sklearn}\href{https://scikit-learn.org/stable/modules/classes.html\#module-sklearn.datasets}{https://scikit-learn.org/stable/modules/classes.html\#module-sklearn.datasets}} with OoD samples and extend the model by three novel classes. Details on this experiment are provided in the appendix. The following method description is also illustrated in~\cref{fig:method_overview}.

% \paragraph{Motivating Example}

\paragraph{I) Learning Model}

For an input image $x\in\mathcal{X}$, let $f(x)\in(0,1)^q$ denote the softmax probabilities of some image classification model $f:\mathcal{X} \to (0,1)^q$ with underlying classes $\mathcal{C}=\{1,\ldots,q\}$. Consider a test dataset which includes images from classes $c\in\{1,\ldots,q, q+1,\ldots\}$. Note that our framework does not necessarily assume labels for the test data as these will be only used for evaluation and not during the training. Furthermore, let $u(f(x))\in [0,1]$ denote some arbitrary uncertainty score which derives from the predicted class-probabilities $f(x)$. Thus, a test image $x$ is considered to be OoD, if $u(f(x))>\tau$ for some threshold $\tau\in[0,1]$. 
% \hg{We then write $x\in \mathcal{X}^\mathrm{OoD}$.} \hg{HG: Die Notation $\mathcal{X}$ und $\mathcal{X}^\mathrm{OoD}$ ist nicht ganz glatt, $\mathcal{X}^\mathrm{OoD}$ wäre konsequenter?!}

Next, we extend the initial model $f$ by $k\in\mathbb{N}$ empty classes in the final classification layer, which is then denoted as $f^k:\mathcal{X} \to (0,1)^{q+k}$, and fine-tune it on the OoD data $\mathcal{X}^\mathrm{OoD}$. Therefore, we compute pairwise distances $d_{ij} = d(x_i,x_j)$ for all $(x_i,x_j)\in\mathcal{X}^\mathrm{OoD}\times\mathcal{X}^\mathrm{OoD}$ as a pre-processing step, e.g.\  using the pixel-wise Euclidean distance or any distance metric in the feature space of some embedding network. 
The model $f^k$ is then fine-tuned on (a subset of) the initial training data $\mathcal{X}^\mathrm{train}$, enriched with the OoD samples from the test data. For the in-distribution samples $(x,y)$, we compute the cross-entropy loss 
\begin{equation}
    \ell_\mathrm{ce}(x, y) = - \sum_{c=1}^q \mathbbm{1}_{\{c=y\}} \log(f^k_c(x)) \; .\label{eq:ce_loss}
\end{equation}
Further, we entice the model to predict one of the empty classes $q+1,\ldots,q+k$ for OoD data by minimizing the class-probabilities $f^k_1(x),\ldots,f^k_q(x)$, $x\in\mathcal{X}^\mathrm{OoD}$, i.e.\ , by computing
\begin{equation}
    \ell_\mathrm{ext}(x) = \frac{1}{q} \sum_{c=1}^q f^k_c(x) \; .\label{eq:extension_loss}
\end{equation}
Finally, we aim to divide the data among the empty classes based on their 
% \outhg{visual} 
similarity. Thus, our clustering loss is computed pair-wise as 
\begin{equation}
    \ell_\mathrm{cluster}(x_i,x_j) = \frac{\alpha}{q+k} \cdot d_{ij}\cdot \sum_{c=1}^{q+k} f^k_c(x_i) f^k_c(x_j)  \; ,\label{eq:cluster_loss}
\end{equation}
where $\alpha\in\mathbb{R}_{>0}$ can be adjusted to control the impact of the clustering loss function. 
Together, these three loss functions give the overall objective
\begin{align}\label{eq:loss}
    L &= \lambda_1 \mathbb{E}_{(x,y)\sim\mathcal{X}^\mathrm{train}}[\ell_\mathrm{ce}(x,y)] \nonumber \\
    &+ \lambda_2 \mathbb{E}_{x\sim\mathcal{X}^\mathrm{OoD}}[\ell_\mathrm{ext}(x)] \\
    &+ \lambda_3 \mathbb{E}_{x_i, x_j\sim\mathcal{X}^\mathrm{OoD}}[\ell_\mathrm{cluster}(x_i,x_j)] \nonumber \; ,
\end{align}
where the hyperparameters $\lambda_1$, $\lambda_2$ and $\lambda_3$ can be adjusted to balance the impact of the objectives.

% \paragraph{II) \outjl{Methods for} OoD Detection}
\paragraph{II) OoD Detection}

OoD detection is a pre-processing part of our framework, which can be exchanged in a plug and play manner. In our experiments, we implemented entropy maximization~\cite{Hendrycks2018DeepAD} for image classification and thus perform OoD detection by thresholding on the softmax entropy.

The idea of entropy maximization is the inclusion of \emph{known unknowns} into the training data of the initial model in order to entice it to exhibit a high softmax entropy 
\begin{equation}
    u(x) = -\frac{1}{\log(q)} \sum_{c=1}^q f_c(x) \log(f_c(x))
\end{equation}
on OoD data $x\in\mathcal{X}^\mathrm{OoD}$. %\hg{HG: $\mathcal{X}^\mathrm{OoD}$?!}
Therefore, during training the initial model, we compute the entropy maximization loss
\begin{equation}
        \ell_\mathrm{em}(x) = - \sum\limits_{c=1}^q \frac{1}{q} \log(f_c(x))
\end{equation}
for known unknowns $x\in \mathcal{X}^\mathrm{OoD}$, giving the overall objective
\begin{align}
    L &= \lambda~ \mathbb{E}_{(x,y)\sim\mathcal{X}^\mathrm{train}}[\ell_\mathrm{ce}(x,y)] \nonumber \\
    &+ (1-\lambda) ~\mathbb{E}_{x\sim\mathcal{X}^\mathrm{OoD}}[\ell_\mathrm{em}(x)] \; .
\end{align}
In the Two Moons example, these OoD data was uniformly distributed noise. For image classification, we employ the domain-agnostic data augmentation technique mixup~\cite{Zhang2017mixupBE}. 
This is, an OoD image is obtained by computing the average of two in-distribution samples. Entropy maximization was also introduced for semantic segmentation of street scenes~\cite{Jourdan2019IdentificationOU,Chan_2021_ICCV}, where the OoD samples originate from the COCO dataset~\cite{Lin2014MicrosoftCC}. Furthermore, the OoD loss and data was only included in the final training epochs, which means that existing networks can be fine-tuned for entropy maximization.

% We employ entropy maximization for image classification and thus perform OoD detection by thresholding on the softmax entropy, i.e.\ , $u(x) = H_f(x),~x\in\mathcal{X}$. 

% \paragraph{III) \outjl{Computation of the} Distance Matrix}
\paragraph{III) Distance Matrix}

Next, we compute pair-wise distances for the detected OoD samples, which constitute the OoD dataset for the incremental learning. For simple datasets such as TwoMoons or MNIST, the distance can be measured directly between the data samples. For MNIST, this is done by flattening the images and computing the Euclidean distance between the resulting vectors. For more complex datasets, we employ embedding networks to extract useful features of the images. These embedding networks are arbitrary image classification models, trained on large datasets such as ImageNet~\cite{ImageNet} or CIFAR100~\cite{Krizhevsky2009LearningML}, which need to be chosen carefully and individually for each experiment as the clustering loss strongly depends on their ability to extract separable features for the known and especially the novel classes.

% \outhg{Afterwards, t}\hg{T}he distances \outhg{could}  either \outhg{be directly} \hg{are}  computed in the high-dimensional feature space \hg{directly}, or, for the sake of transparency and \outhg{visualizability} \hg{better visual control}, in a low-dimensional \outhg{projection} \hg{re-arrangement}. \outhg{By a}\hg{A}pplying the manifold learning technique UMAP~\cite{McInnes2018UMAPUM} \outhg{on}to the entire test data, we reduce the dimension of the feature space to two
% % , computing the distance in the input space with the cosine metric
% . \outhg{Then, t}\hg{T}he distance matrix is \hg{then} \outhg{constructed by computing} \hg{computed as} the Euclidean distances in the low-dimensional space for all pairs of OoD samples.

The feature distances are either computed in the high-dimensional feature space directly, or, for the sake of transparency and better visual control, in a low-dimensional re-arrangement. Applying the manifold learning technique UMAP~\cite{McInnes2018UMAPUM} to the entire test data, we reduce the dimension of the feature space to two
% , computing the distance in the input space with the cosine metric
. The distance matrix is then computed as the Euclidean distances in the low-dimensional space for all pairs of OoD samples.

% \paragraph{IV) \outjl{Loss Functions for} Incremental Learning}
\paragraph{IV) Incremental Learning}

For class-incremental learning, we minimize three different loss functions defined in \cref{eq:ce_loss,eq:extension_loss,eq:cluster_loss}. The cross-entropy loss (\ref{eq:ce_loss}) is computed for in-distribution to mitigate catastrophic forgetting~\cite{McCloskey1989CatastrophicII}. The OoD samples are pushed towards the novel classes by the extension loss (\ref{eq:extension_loss}), which is minimized whenever the probability mass is concentrated in the empty classes, i.e.\ ,
\begin{equation}
    \ell_\mathrm{ext}(x) \to 0 \text{ for } \sum_{c=q+1}^{q+k} f^k_c(x) \to 1,~ x\in\mathcal{X}^\mathrm{OoD} \; .
\end{equation}
The cluster loss (\ref{eq:cluster_loss}) is computed for all pairs of OoD candidates contained in a batch. Thus, it has a runtime complexity of $\mathcal{O}(n^2)$, as for $n$ OoD candidates, we need to compute $\frac{n^2 - n}{2}$ terms. Furthermore, the minimum of the cluster loss is probably greater than zero, as samples which belong to the same class rarely share exactly the same features. To reach this minimum for two OoD samples $x_i,x_j$ with a large distance, they should be assigned to different classes, i.e.\ , whenever $f^k_c(x_i)$ is significantly different from zero, we desire that $f^k_c(x_j)$ becomes small.

\section{Adjustments for Semantic Segmentation}\label{sec:semseg}

\begin{figure}[t]
    \captionsetup[subfloat]{labelformat=empty}
    \centering
    \subfloat[]{\includegraphics[width=0.23\textwidth]{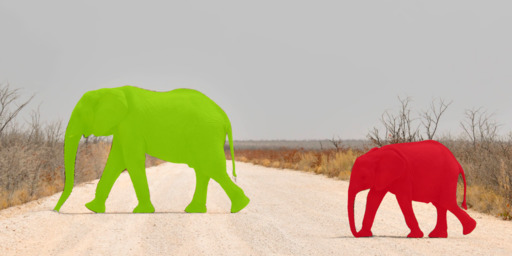}}~
    \subfloat[]{\includegraphics[width=0.23\textwidth]{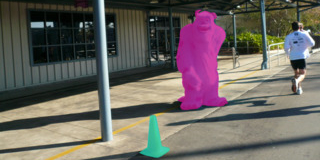}}
    \caption{In semantic segmentation, each of the OoD objects is assigned a unique ID, no matter if they belong to the same novel class as the elephants, or to different classes as the cone and the monster costume~\cite{chan2021segmentmeifyoucan}.}
    \label{fig:semseg_OoD}
\end{figure}

Let $H\times W$ denote the resolution of the images $x\in\mathcal{X}$. Then, the softmax output of a semantic segmentation DNN $f:\mathcal{X}\to(0,1)^{H \times W \times q}$ provides class-probabilities for image pixels, denoted as $z=(h,w)\in\mathcal{Z}$. Thus, the OoD detector must not only identify OoD images, but also give information about their pixel positions. To store these information, we generate OoD instance masks as illustrated in \cref{fig:semseg_OoD} by thresholding on the obtained OoD score and by distinguishing between connected components in the resulting OoD mask. 

% In addition to the cross-entropy loss, we compute a distillation loss~\cite{Michieli2021KnowledgeDF}
% \begin{equation}
%     \ell_\mathrm{d}(x) = -\frac{1}{|\mathcal{Z}|} \sum_{z\in\mathcal{Z}}\frac{1}{q}\sum_{c=1}^q f_{z,c}(x)\log(f^k_{z,c}(x))
% \end{equation}
% for in-distribution data to retain the performance on the previously-known classes. 
For semantic segmentation, the loss functions are computed for pixels of OoD objects instead of images. Let $\mathcal{Z}_s$ denote the set of pixel positions which belong to an OoD candidate $s\subseteq x$.
The extension loss is computed equivalently to \cref{eq:extension_loss} as
\begin{equation}
    \ell_\mathrm{ext}(s) = -\frac{1}{|\mathcal{Z}_s|} \sum_{z\in\mathcal{Z}_s}\frac{1}{q}\sum_{c=1}^q f^k_{z,c}(x) \; .\label{eq:semseg_extension_loss}
\end{equation}
For two OoD candidates $s_i\subseteq x_i,s_j\subseteq x_j$ with distance $d_{ij}$, the cluster loss is computed as
\begin{equation}
    \ell_\mathrm{cluster}(s_i,s_j) = \frac{\alpha}{q+k} d_{ij} \sum_{c=1}^{q+k} \overline{f^k_c(x_i)}~\overline{f^k_c(x_j)}  \; ,\label{eq:semseg_cluster_loss}
\end{equation}
where 
\begin{equation}
    \overline{f^k_c(x)} = \frac{1}{|\mathcal{Z}_{s}|}\sum_{z\in \mathcal{Z}_{s}}f^k_{z,c}(x)
\end{equation}
denotes the mean softmax probability over all pixels $z\in\mathcal{Z}_{s}$ for some class $c\in\{1,\ldots,q+k\}$.

For OoD detection in semantic segmentation, we adapt a meta regression approach~\cite{rottmann2019prediction,rottmann2019uncertainty}, using uncertainty measures such as the softmax entropy and further information which derives from the initial model's output, to estimate the prediction quality on a segment-level. Here, a segment denotes a connected component in the semantic segmentation mask, which is predicted by the initial model. That is, meta regression is a post-processing approach to quantify uncertainty aggregated over segments, and considering that the model likely is highly uncertain if confronted with OoD objects, it can be applied for OoD detection. In contrast to image classification, where images are either OoD or not, semantic segmentation is performed on images which can contain in-distribution and OoD pixels at the same time. Aggregating uncertainty scores across segments simplifies the detection of OoD objects as contiguous OoD pixels, since it removes the high uncertainty for class boundaries.

For an initial DNN, we use the training data to fit a gradient boosting model as meta regressor, which then estimates segment-wise uncertainty scores $u(s)$ for all segments $s\subseteq x\in\mathcal{X}$.

\section{Numerical Experiments}

We perform several experiments for image classification on MNIST~\cite{LeCun1998GradientbasedLA}, FashionMNIST~\cite{Xiao2017FashionMNISTAN}, CIFAR10~\cite{Krizhevsky2009LearningML} and Animals10, as well as on Cityscapes~\cite{Cordts2016TheCD} to evaluate our method for semantic segmentation. To this end, we extend the initial models by empty classes, i.e.\ , neurons in the final classification layer with randomly initialized weights, and fine-tune them on OoD data, retraining with fixed encoder. For evaluation, we provide accuracy scores \-- separately for known and novel classes \-- for image classification, (mean) Intersection over Union (IoU), precision and recall values for semantic segmentation. 

The OoD classes in the following experiments were all chosen in a way that they are semantically far away from each other. For example, the Animals10 classes \emph{horse} $(1)$, \emph{cow} $(6)$ and \emph{sheep} $(7)$ are semantically related, as they are all big animals which are mostly on the pasture, whereas \emph{elephant} $(2)$ and \emph{spider} $(8)$ are well separable classes, which is also visible in the two-dimensional feature space. However, we will also provide evaluation metrics averaged over multiple runs with randomly picked OoD classes in the appendix.

\subsection{Experimental Setup}

For each experiment, we consider the following dataset splits: the \emph{training data} denotes images with ground truth for the initially known classes. We train the initial model on these images and replay them during the training of the extended model to avoid catastrophic forgetting. The \emph{test data} consist of unlabeled images which include both, known and unknown classes. This dataset is fed into the OoD detector to identify \emph{OoD data}, on which the model gets extended. The \emph{evaluation dataset} includes images with ground truth for known and novel classes and is used to evaluate the models. If there are such labels available for the test data, evaluation images may be the same as the test images.

Our approach requires prior OoD detection. Here, we only provide the experimental setup for fine-tuning the extended model. For all experiments, we tuned the weighting parameters $\lambda_1,\lambda_2,\lambda_3$ in~\cref{eq:loss} by observing all loss functions separately over several epochs using different parameter configurations to ensure that each loss term decreases. The following descriptions of the experiments, sorted by the datasets, include the network architecture, the known and novel classes, information about the dataset splits and the generation of the distance matrix. For further information about the experiments, which also includes the TwoMoons experiment, we refer to the appendix.

\paragraph{MNIST}
We employ a shallow neural network consisting of two convolutional layers, each followed by a ReLU activation function and max pooling, and a fully connected layer. From the digits $0,\ldots,9$, we select $0,5$ and $7$ as novel classes. All images in the MNIST training set which belong to these classes are excluded from our training data. The MNIST test images compose our test set, and together with the original labels, our evaluation set. The distance matrix is computed as pixel-wise Euclidean distance between the OoD images. 
% We extend the model by $3$ empty classes and fine-tune it for $30$ epochs on the OoD data enriched training split, using the Adam optimizer with learning rate $1e^{-2}$, a batch size of $2500$ and $\alpha=5$ in~\cref{eq:cluster_loss}.

\begin{figure}[t]
    \centering
    \includegraphics[width=0.47\textwidth]{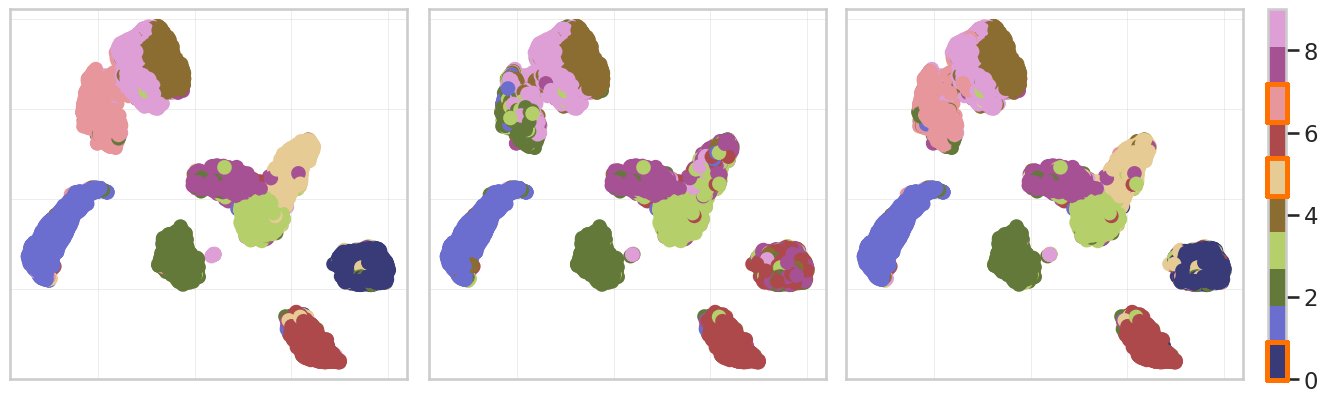}
    \caption{Visualized ground truth \emph{(left)} and prediction of the MNIST dataset by the initial \emph{(middle)} and extended \emph{(right)} model. The three novel classes $0,5$ and $7$ are outlined in orange. The extended model's accuracy is $\sim 94\%$. }\label{fig:result_MNIST}
    
    \includegraphics[width=0.47\textwidth]{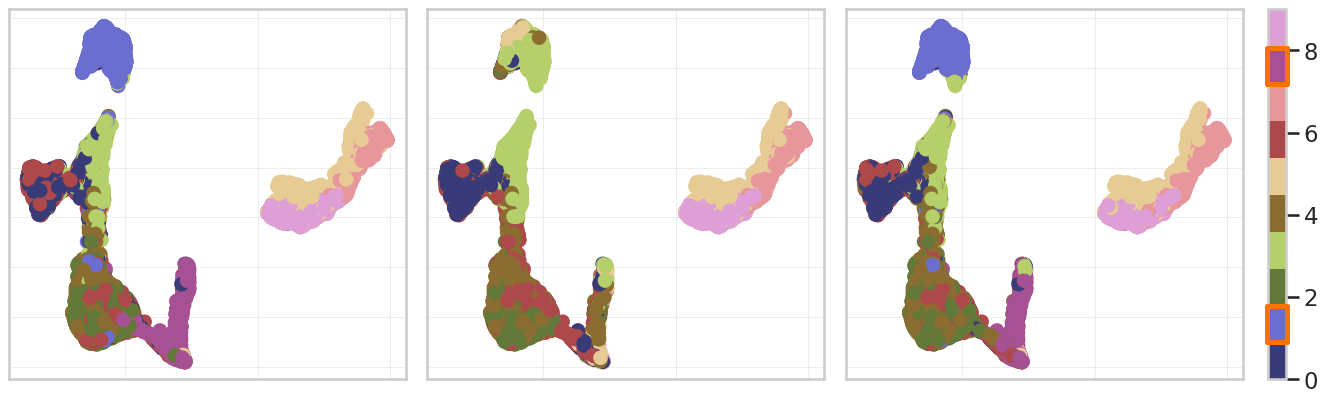}
    \caption{Visualized ground truth \emph{(left)} and prediction of the FashionMNIST dataset by the initial \emph{(middle)} and extended \emph{(right)} model. The two novel classes $1$ and $8$ are outlined in orange. The extended model's accuracy is $\sim 85\%$. }\label{fig:result_FashionMNIST}
\end{figure}

\paragraph{FashionMNIST}
Using the same network architecture as for MNIST, our initial model is trained on eight out of ten classes, excluding the classes \emph{trouser} (1) and \emph{bag} (8). Our dataset splits are created analogously to those from MNIST. Also the distance matrix is obtained analogously by computing the pixel-wise Euclidean distances between the OoD images. 
% We extend the model by $2$ empty classes and fine-tune it for $30$ epochs on the OoD data enriched training split, using the SGD optimizer with learning rate $1e^{-2}$, a batch size of $500$ and $\alpha=2.5$.

\paragraph{CIFAR10}
The setting for CIFAR10 differs slightly from the other experiments to ensure comparability with existing approaches. Thus, as initial model, we employ a ResNet18 which is trained on the whole CIFAR10 training split, including all ten classes. For testing, we enrich the CIFAR10 test split with images from CIFAR100. Therefore, we split CIFAR100 into an unlabeled and a labeled subset: the classes $\{0,\ldots,49\}$ are possible OoD candidates, thus, all samples belonging to these classes are considered to be unlabeled. We extend the CIFAR10 test data by the classes \emph{apple} (0) and \emph{clock} (22), mapping them onto the labels (10) and (11), respectively. As before, we evaluate our models on the labeled test data. The labeled CIFAR100 subset includes the classes $\{50,\ldots,99\}$ and is used together with the CIFAR10 training data to train a ResNet18 as an embedding network. To compute the distances, we feed the whole test data into this embedding network and extract the features of the penultimate layer. These are further projected into a 2D space with UMAP. Then, the distance matrix is computed as the pixel-wise Euclidean distance between the 2D representations of the OoD images. 
% We extend the model by $2$ empty classes and fine-tune it for $30$ epochs on the OoD data enriched training split, using the SGD optimizer with learning rate $1e^{-2}$, momentum $0.9$ and weight decay $0.1$. Further, we use a batch size of $1000$ and $\alpha=5$.

\begin{figure}[t]
    \centering
    \includegraphics[width=0.47\textwidth]{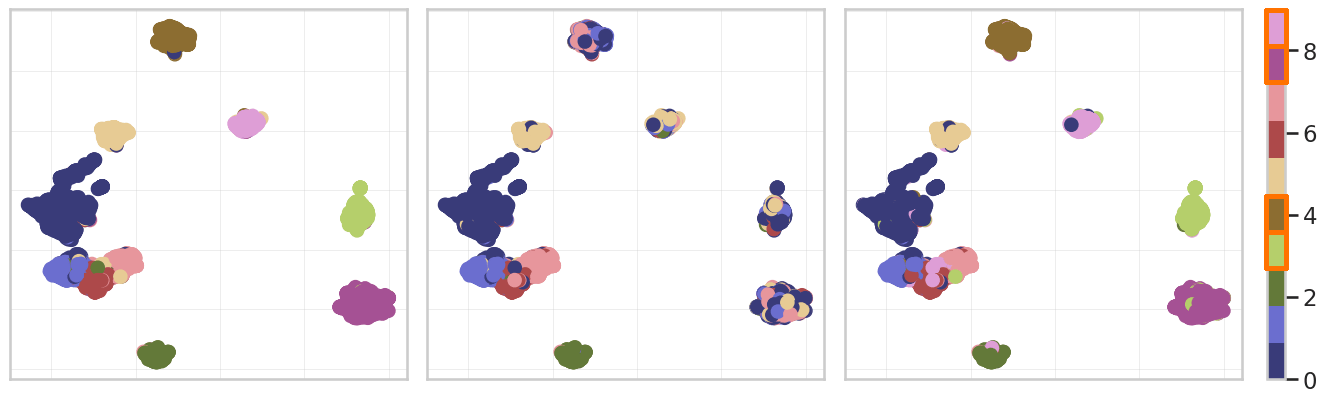}
    \caption{Visualized ground truth \emph{(left)} and prediction of the Animals10 dataset by the initial \emph{(middle)} and extended \emph{(right)} model. The four novel classes $3,4,8$ and $9$ are outlined in orange. The extended model's accuracy is $\sim 95\%$. }\label{fig:result_Animals}
    
    \includegraphics[width=0.47\textwidth]{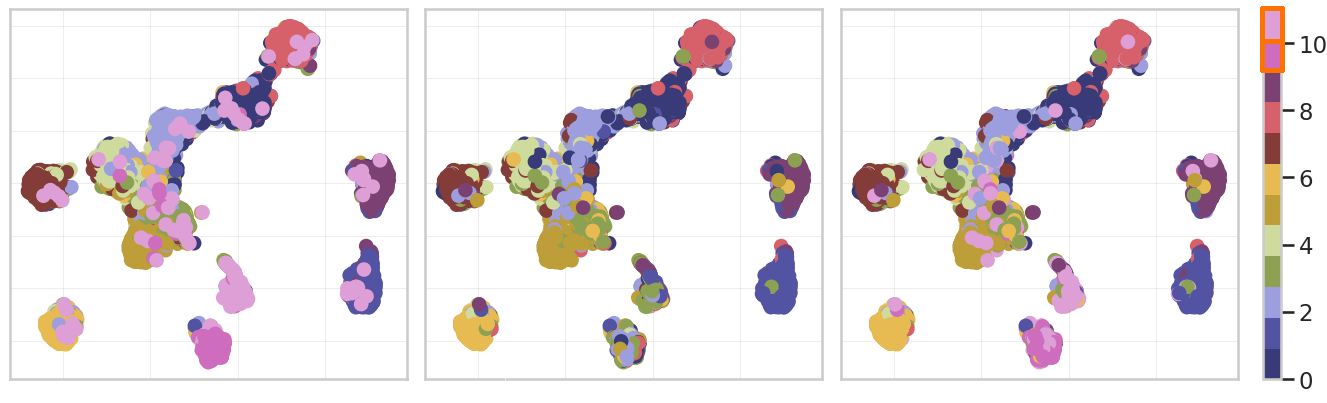}
    \caption{Visualized ground truth \emph{(left)} and prediction of the CIFAR10 dataset by the initial \emph{(middle)} and extended \emph{(right)} model. The two novel classes $10$ and $11$ are outlined in orange. The extended model's accuracy is $\sim 89\%$. }\label{fig:result_CIFAR}
\end{figure}

\begin{table*}[t]
    \centering
    \begin{tabular}{l|l|l||c|c||c|c||c|c}
        \hline
        \multicolumn{9}{c}{\textbf{Image Classification}}\\\hline
        \multicolumn{3}{c||}{} & \multicolumn{2}{c||}{supervised} & \multicolumn{2}{c||}{unsupervised} & \multicolumn{2}{c}{ablation studies}\\\hline
        \textbf{dataset} & \textbf{OoD} & \textbf{accuracy} & \textbf{initial} & \textbf{oracle} & \textbf{ours} & \textbf{baseline} & \textbf{$-$detection} & \textbf{$--$distance} \\\hline\hline
        \multirow{2}{*}{MNIST} & \multirow{2}{*}{$0~~5~~7$} & known & $96.68\%$ & $98.54\%$ & \cellcolor{gray!25}$\mathbf{96.20\%}$ & $95.94\%$ & $97.45\%$ & $96.54\%$ \\
        & & novel & - & $95.85\%$ & \cellcolor{gray!25}$\mathbf{97.94\%}$ & $84.62\%$ & $74.52\%$ & $97.00\%$  \\\hline
        \multirow{2}{*}{FashionMNIST} & \multirow{2}{*}{$1~~8$} & known & $81.54\%$ & $83.75\%$ & \cellcolor{gray!25}$81.41\%$ & $\mathbf{85.08\%}$ & $81.89\%$ & $81.39\%$\\
        & & novel & - & $90.83\%$ & \cellcolor{gray!25}$90.05\%$ & $\mathbf{92.85\%}$ & $89.90\%$ & $95.00\%$ \\\hline
        \multirow{2}{*}{CIFAR10} & \multirow{2}{*}{$10~~11$} & known & $91.45\%$ & $91.86\%$ & \cellcolor{gray!25}$\mathbf{90.51\%}$ & $90.29\%$ & $88.90\%$ & $86.94\%$\\
        & & novel & - & $89.53\%$ & \cellcolor{gray!25}$\mathbf{70.00\%}$ & $33.40\%$ & $78.80\%$ & $87.00\%$ \\\hline
        \multirow{2}{*}{Animals10} & \multirow{2}{*}{$3~~4~~8~~9$} & known & $96.29\%$ & $95.80\%$ & \cellcolor{gray!25}$\mathbf{93.76\%}$ & $92.78\%$ & $94.46\%$ & $95.20\%$ \\
        & & novel & - & $97.65\%$ & \cellcolor{gray!25}$\mathbf{96.68\%}$ & $72.59\%$ & $97.02\%$ & $97.90\%$ \\\hline
    \end{tabular}
    \caption{Quantitative evaluation of the image classification experiments. For all evaluated models, the accuracy is stated separately for the previously-known and the unlabeled novel classes. The highest scores for the unsupervised approaches are bolded.}
    \label{tab:results_ic}
\end{table*}

\paragraph{Animals10}
As initial model, we employ a ResNet18 which is trained on six out of ten classes. As novel classes we selected \emph{butterfly} (3), \emph{chicken} (4), \emph{spider} (8) and \emph{squirrel} (9). The dataset splits are obtained analogously to those from MNIST. The distances are computed as for CIFAR10, but employing a DenseNet201, which is trained on ImageNet with $1,\!000$ classes, as embedding network. 
% We extend the model by $4$ empty classes and fine-tune it for $30$ epochs on the OoD data enriched training split, using the Adam optimizer with learning rate $5e^{-3}$, a batch size of $1000$ and $\alpha=2.5$.

\paragraph{Cityscapes}
For comparison reasons with the baseline, we adapt the experimental setup from \cite{uhlemeyer2022towards}, where the class labels \emph{human (person, rider), car} and \emph{bus} are excluded from the $19$ Cityscapes evaluation classes. Like the baseline, we extend the DNN by two empty classes and exclude the class \emph{bus} from the evaluation. Thus, we train a semantic segmentation DeepLabV3+ with WideResNet38 backbone on $2,\!500$ training samples with $15$ trainable classes. We apply meta regression to the Cityscapes test data and crop out image patches tailored to the predicted OoD segments, i.e.\ , connected component of OoD pixels. Afterwards, we compute distances between these image patches analogously to Animals10 as the Euclidean distances between 2D representations of features which we obtain by feeding the patches into a DenseNet201 trained on $1,\!000$ ImageNet classes. 
% We extend the DNN by two empty classes and fine-tune it for $200$ epochs on a subset of the training data enriched with images containing OoD regions, using the Adam optimizer with learning rate $5e^{-3}$, a batch size of $10$ and $\alpha=2.5$.

\subsection{Evaluation \& Ablation Studies}

We compare our evaluation results to the following baselines. For image classification, we employ the k-means clustering algorithm to pseudo-label the OoD data samples and fine-tune the model on the pseudo-labeled data using the cross-entropy loss. For semantic segmentation, we compare with the method presented in \cite{uhlemeyer2022towards}, which also employs clustering algorithms in the embedding space to obtain pseudo-labels. Furthermore, to get an idea of the maximum achievable performance, we train oracle models which have learned all available classes in a fully supervised manner.

For the ablation studies, we evaluate our image classification approach on ``clean'' OoD data ($-$detection). Therefore, we do not detect the OoD samples in the test data by thresholding on some anomaly score, but by considering the ground truth. In this way, we simulate a perfect OoD detector. Since the results of our method are also affected by the quality of the distance matrix, we further analyze our method for a synthetic distance matrix ($--$distance), where two OoD samples $x_i,x_j\in\mathcal{X}^\mathrm{OoD}$ have a distance $d(x_i,x_j)=0$ if they stem from the same class, $d(x_i,x_j)=1$ otherwise. Thus, the OoD samples are labeled by the distance matrix and the fine-tuning is supervised, allowing a pure comparison of our loss functions with the cross-entropy loss. We do not provide ablation studies for semantic segmentation, since the Cityscapes test data does not include publicly available annotations.

\paragraph{Image Classification}
As shown in~\cref{tab:results_ic} and visualized in~\cref{fig:result_MNIST,fig:result_FashionMNIST,fig:result_CIFAR,fig:result_Animals}, our approach exceeds the baseline's accuracy for novel classes by $36.60$ and $24.09$ percentage points (pp) for CIFAR10 and Animals10, respectively. This is mainly caused by in-distribution samples which are false positive OoD predictions, or by OoD samples which are embedded far away from their class centroids. Consequently, different OoD classes are assigned to the same cluster by the k-means algorithm. As our approach uses soft labels, the DNN is more likely to reconsider the choice of the OoD detector during fine-tuning. 

In the ablation studies, we omit the OoD detector ($-$detection) and select the OoD samples based on their ground truth instead. Thereby, we observe an improvement of the accuracy of novel classes for the CIFAR10 and Animals10 datasets, while the performance remains constant for FashionMNIST and significantly decreases for MNIST. We further compute a ground truth distance matrix ($--$distance) with distances $0$ and $1$ for samples belonging to the same or to different classes, respectively. Since this is supervised fine-tuning, these DNNs are comparable to oracles. We observe, that the oracles tend to perform better on the initial and worse on the novel classes. However, this might be a consequence of the class-incremental learning. 

\begin{figure*}[t]
    \captionsetup[subfloat]{labelformat=empty}
    \centering
    \subfloat[]{ \includegraphics[trim={200px 100px 0 0},clip,width=0.24\textwidth]{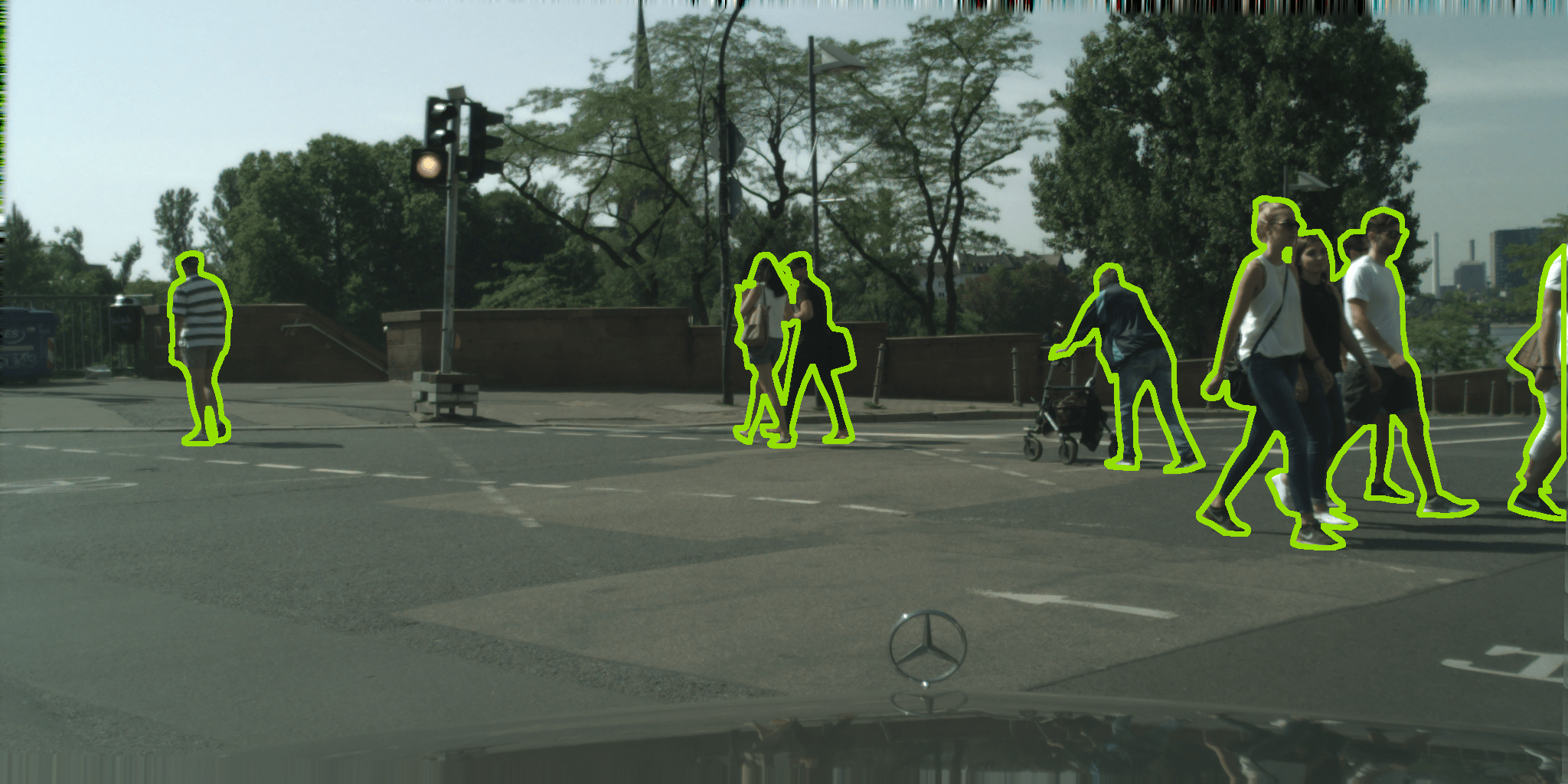}}\hfill
    \subfloat[]{ \includegraphics[trim={200px 100px 0 0},clip,width=0.24\textwidth]{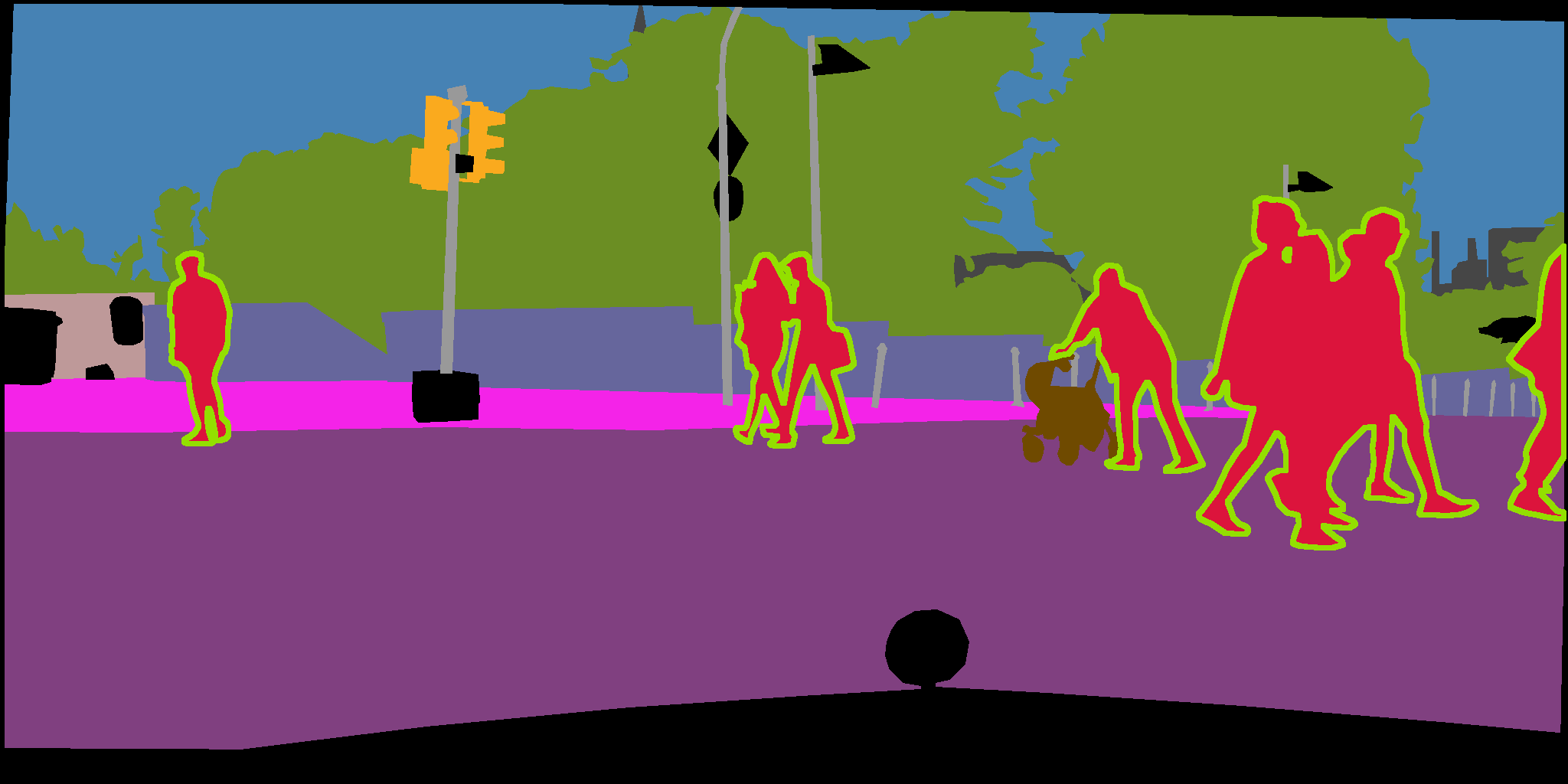}}\hfill
    \subfloat[]{ \includegraphics[trim={200px 100px 0 0},clip,width=0.24\textwidth]{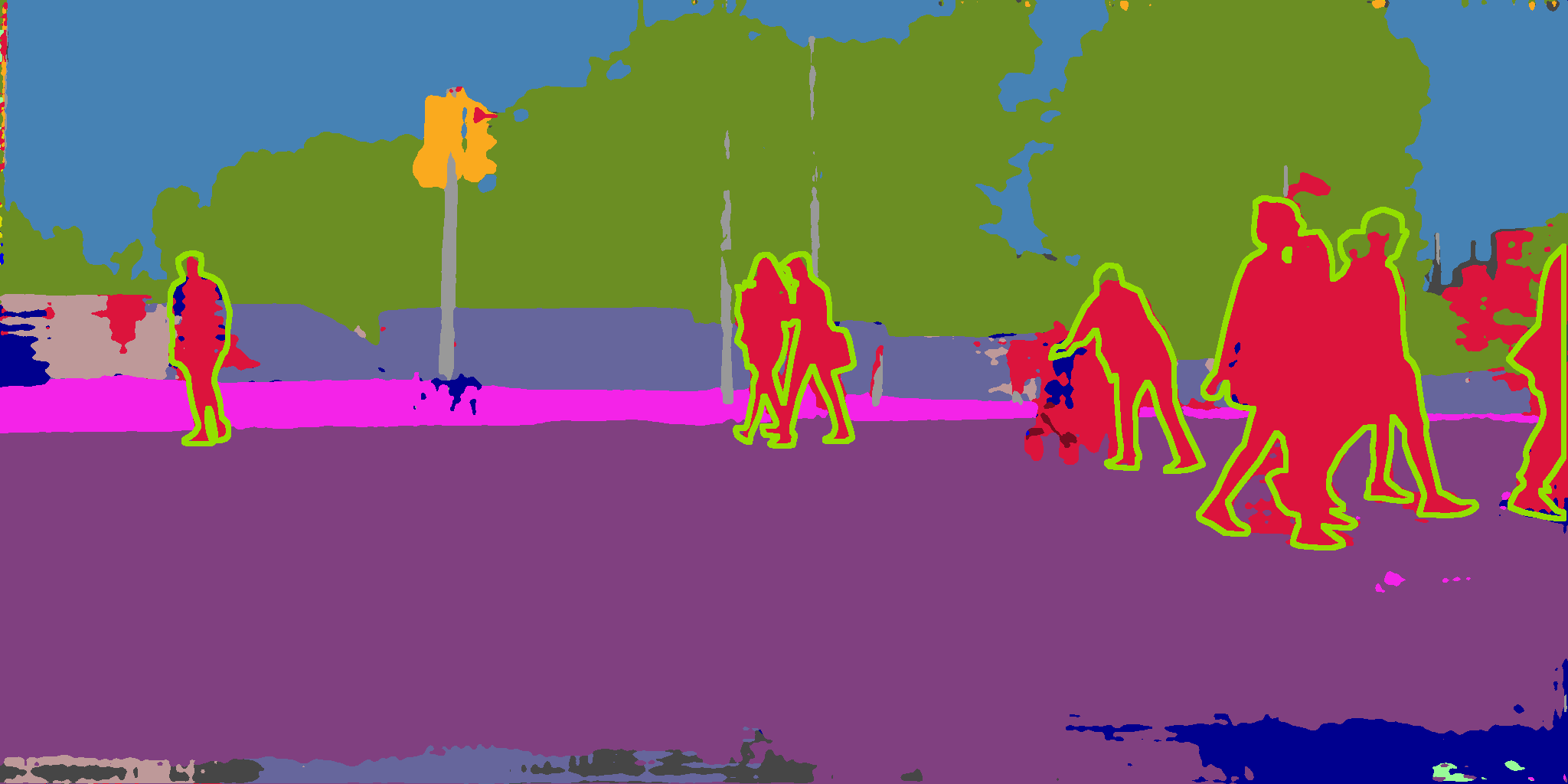}}\hfill
    \subfloat[]{ \includegraphics[trim={200px 100px 0 0},clip,width=0.24\textwidth]{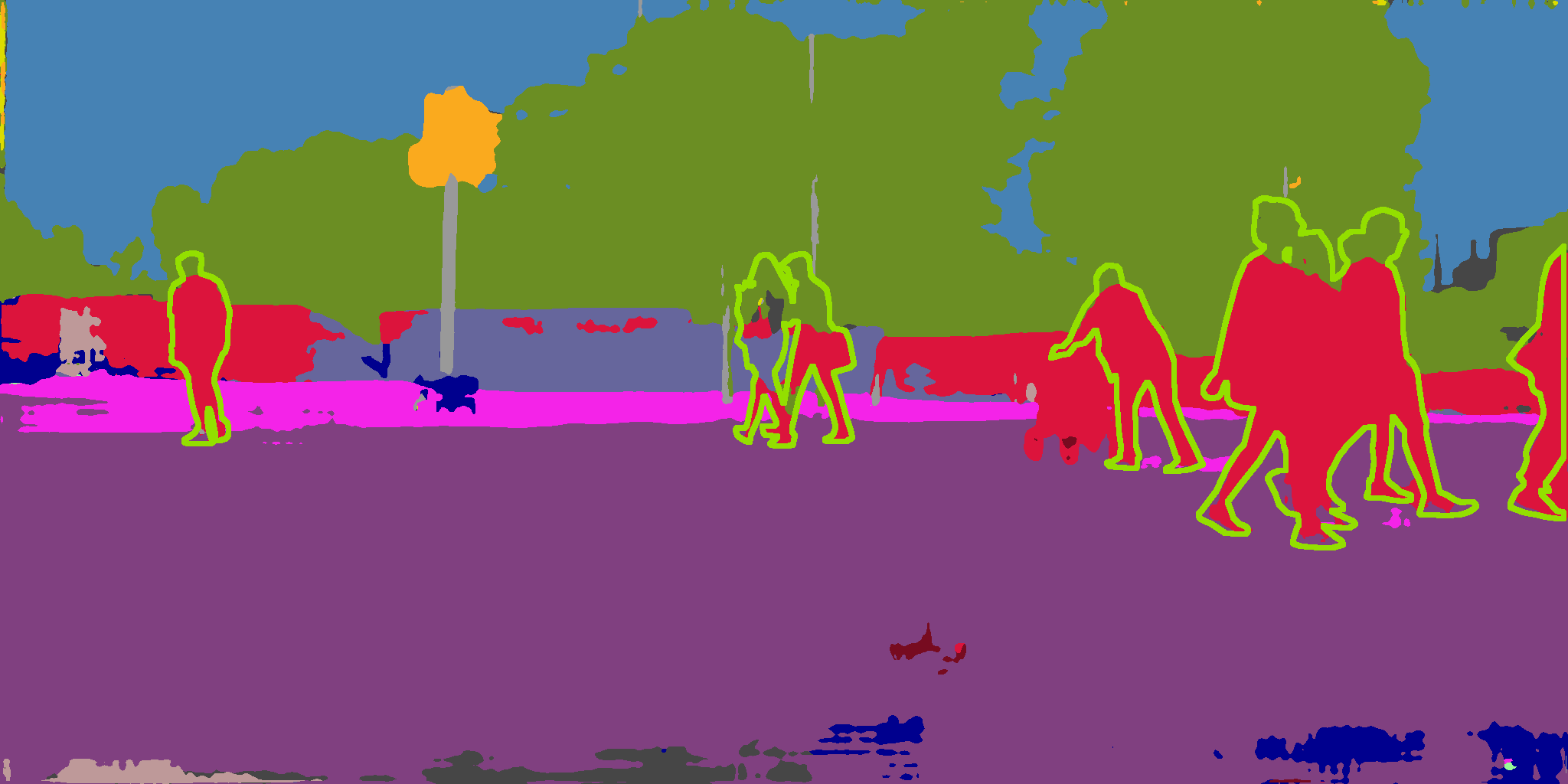}}\\[-2em]
    \subfloat[image]{ \includegraphics[trim={400px 400px 400px 0},clip,width=0.24\textwidth]{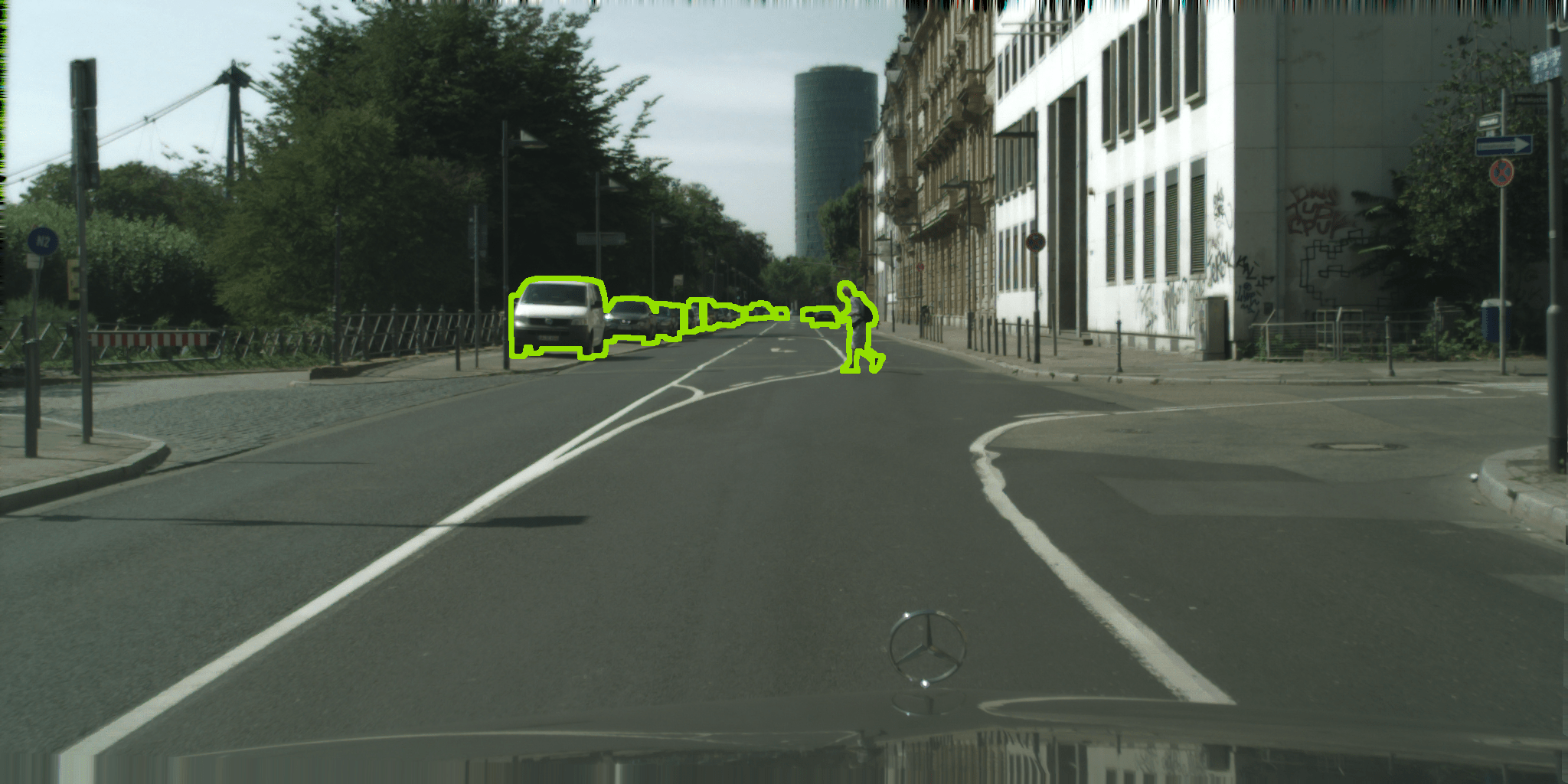}}\hfill
    \subfloat[ground truth]{ \includegraphics[trim={400px 400px 400px 0},clip,width=0.24\textwidth]{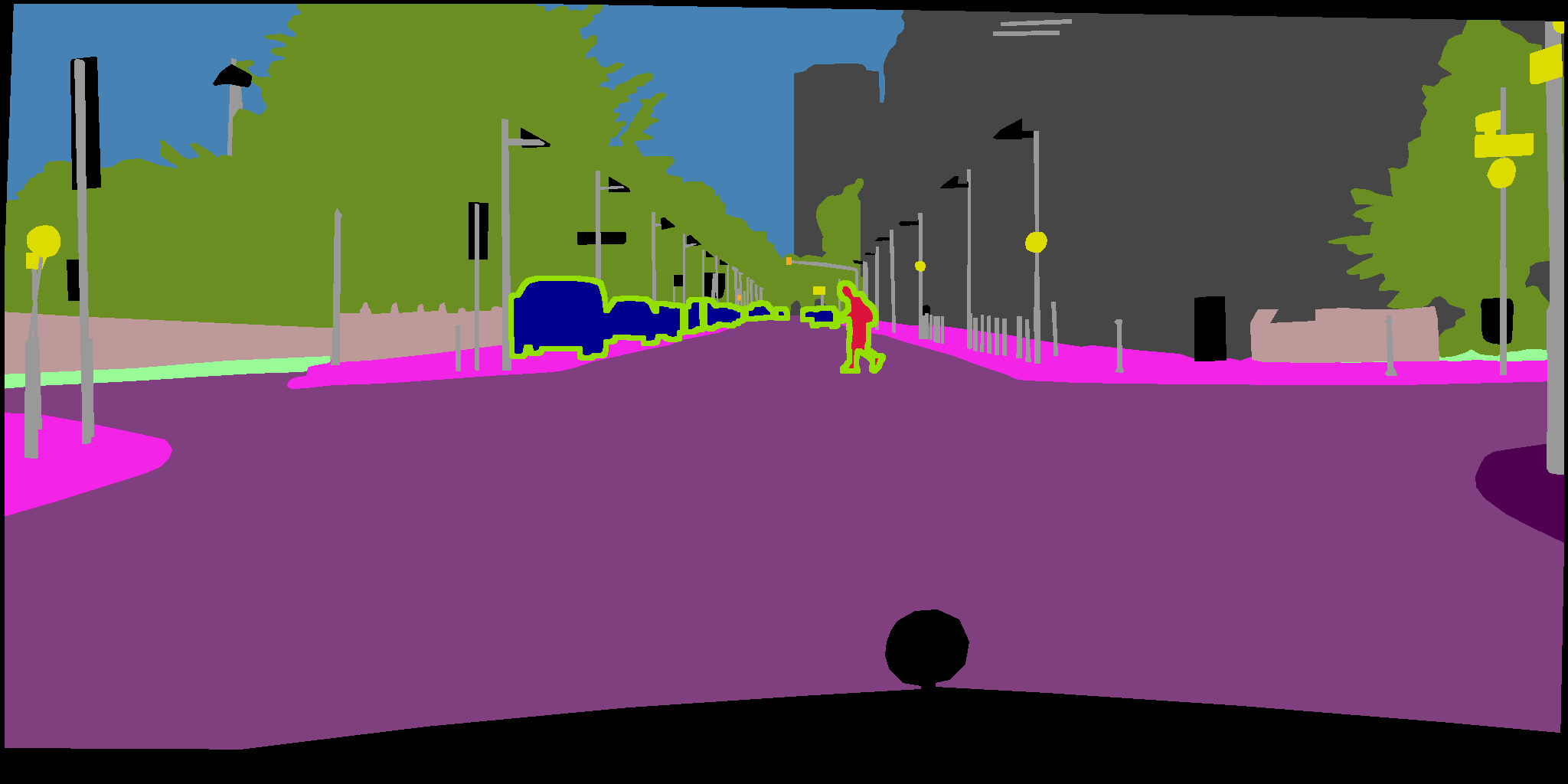}}\hfill
    \subfloat[ours]{ \includegraphics[trim={400px 400px 400px 0},clip,width=0.24\textwidth]{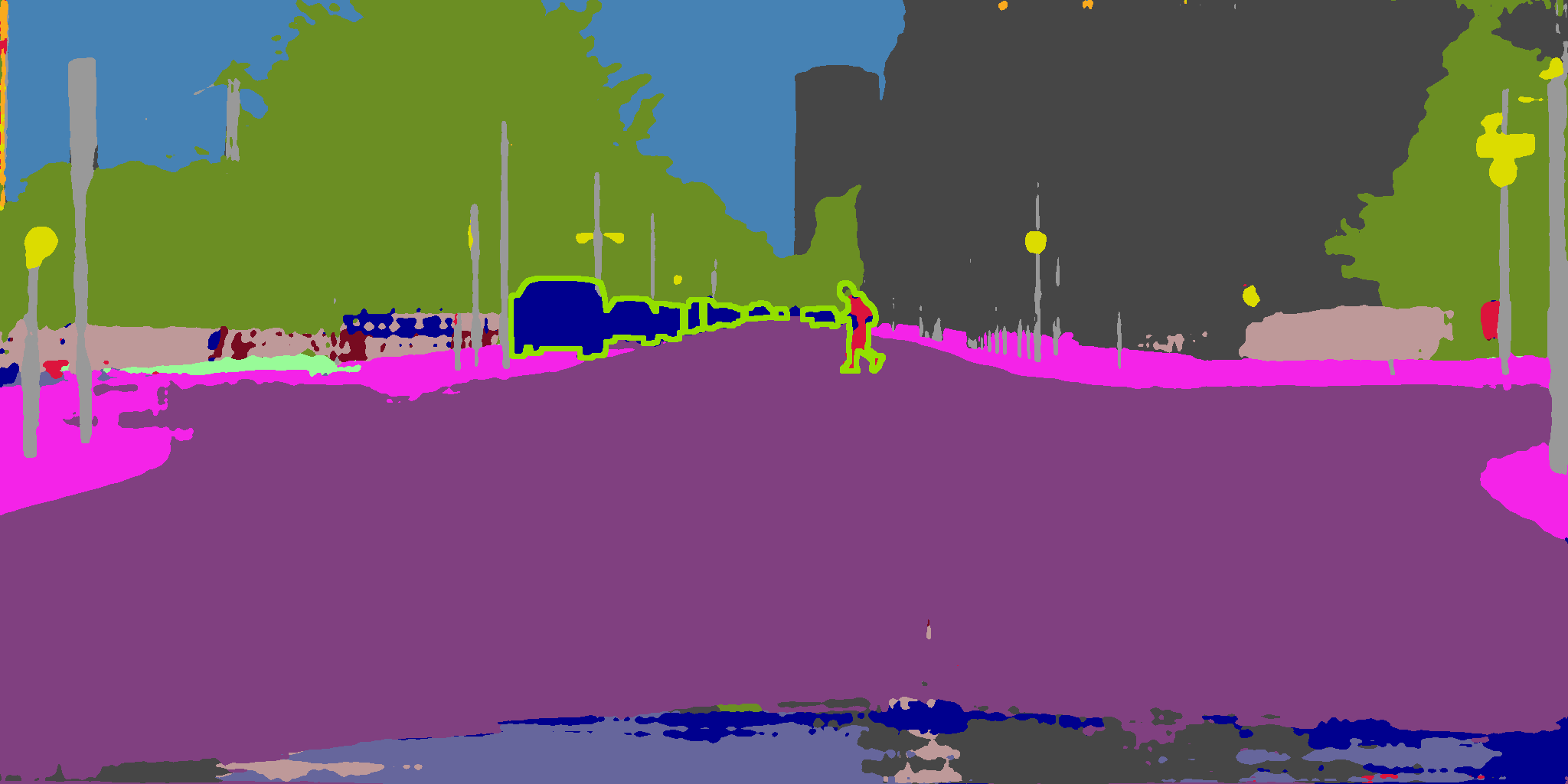}}\hfill
    \subfloat[baseline]{ \includegraphics[trim={400px 400px 400px 0},clip,width=0.24\textwidth]{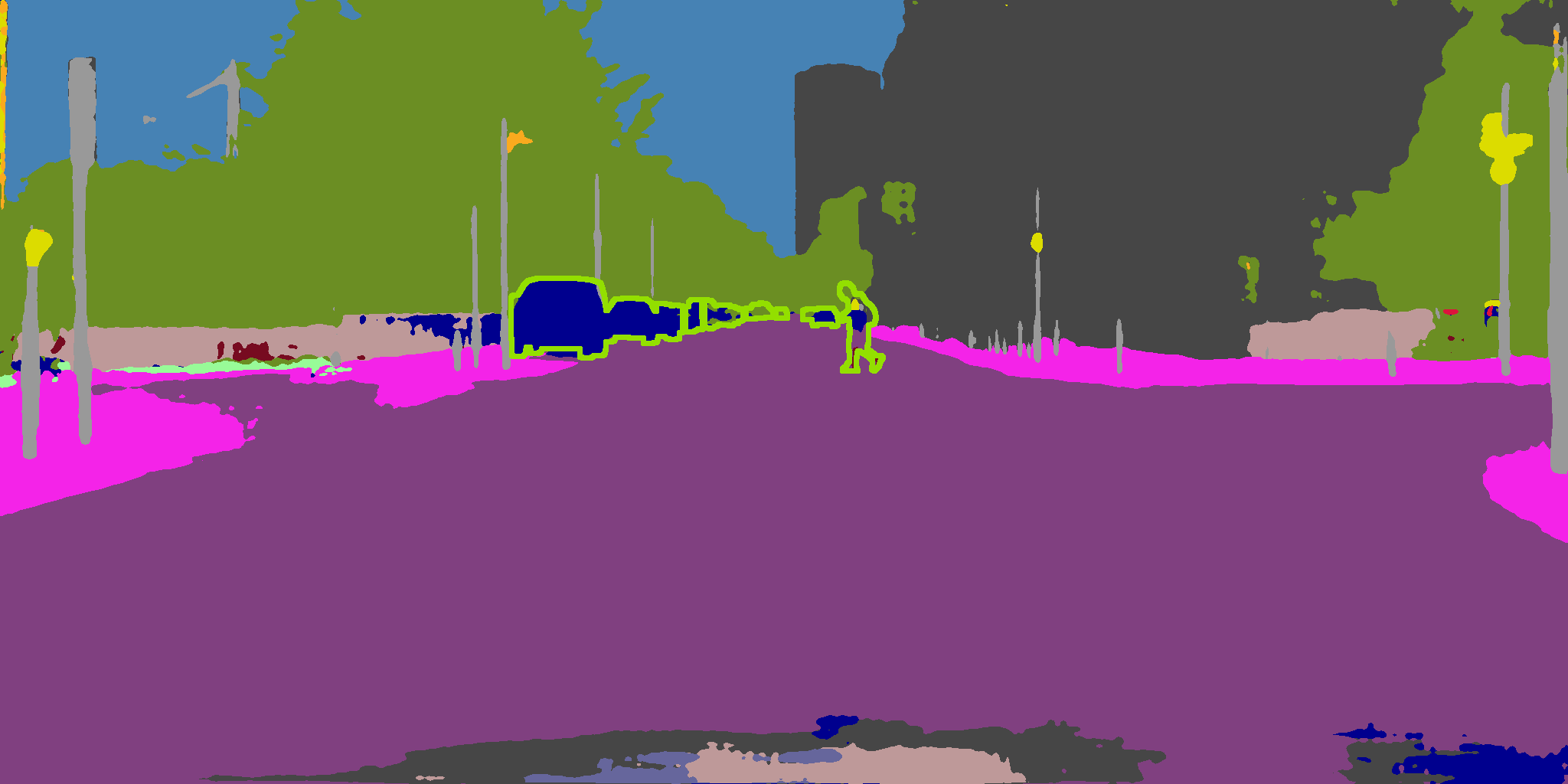}}
    \caption{Visual comparison of the segmentation masks produced by our method and by the baseline for two image cutouts from the Cityscapes validation dataset. The ground truth contours of the novel classes are highlighted with green.}\label{fig:results_cityscapes}
\end{figure*}

\begin{table*}[t]
    \centering
    \begin{tabular}{l|l|l||c|c||c|c}
        \hline
        \multicolumn{7}{c}{\textbf{Semantic Segmentation}}\\\hline\hline
        \multicolumn{3}{c||}{} & \multicolumn{2}{c||}{supervised} & \multicolumn{2}{c}{unsupervised}\\\hline
        \textbf{dataset} & \textbf{class} & \textbf{metric} & \textbf{initial} & \textbf{oracle} &  \textbf{ours} & \textbf{baseline} \\\hline\hline
        \multirow{9}{*}{Cityscapes} & $0,\ldots,14$ & mean IoU & $56.99\%$ & $77.28\%$  & \cellcolor{gray!25} $\mathbf{59.72\%}$ & $57.52\%$\\
        & $15$ (human) & IoU & - & $81.90\%$  & \cellcolor{gray!25}$33.87\%$ & $\mathbf{40.22\%}$ \\
        & $16$ (car) & IoU & - & $94.94\%$  & \cellcolor{gray!25}$\mathbf{84.14\%}$ & $81.27\%$ \\
        \cline{2-7}
        & $0,\ldots,14$ & mean precision & $65.75\%$ & $88.03\%$  & \cellcolor{gray!25}$\mathbf{84.63\%}$ & $78.53\%$ \\
        & $15$ (human) & precision & - & $89.22\%$  & \cellcolor{gray!25}$37.80\%$ & $\mathbf{68.74\%}$ \\
        & $16$ (car) & precision & - & $96.83\%$  & \cellcolor{gray!25}$\mathbf{87.11\%}$ & $86.56\%$ \\
        \cline{2-7}
        & $0,\ldots,14$ & mean recall & $80.88\%$ & $85.38\%$ & \cellcolor{gray!25}$65.38\%$ & $\mathbf{65.78\%}$ \\
        & $15$ (human) & recall & - & $90.90\%$  & \cellcolor{gray!25}$\mathbf{76.54\%}$ & $49.65\%$\\
        & $16$ (car) & recall & - & $97.99\%$  & \cellcolor{gray!25}$\mathbf{96.11\%}$ & $93.05\%$ \\\hline
    \end{tabular}
    \caption{Quantitative evaluation of the semantic segmentation experiment on the Cityscapes dataset. IoU, precision and recall values are provided for both novel classes as well as averaged over the previously-known classes. The highest scores for the unsupervised approaches are bolded.}
    \label{tab:results_ss}
\end{table*}

\paragraph{Semantic Segmentation}
The quantitative results of our semantic segmentation method, reported in~\cref{tab:results_ss}, demonstrate, that the empty classes are ``filled'' with the novel concepts \emph{human} and \emph{car}. Thereby, the performance on the previously-known classes is similar to the baseline even without including a distillation loss~\cite{Michieli2021KnowledgeDF}. For the \emph{car} class, our method outperforms the baseline with respect to IoU ($+2.87$ pp), precision ($+0.55$ pp) and recall ($+3.06$ pp). We lose performance in terms of IoU for the \emph{human} class due to a higher tendency for false positives. However, the false negative rate is significantly reduced, which is indicated by an increase in the recall value of $26.89$ pp. The improved recall score is also visible in~\cref{fig:results_cityscapes}, showing two examples from the Cityscapes validation dataset. In the top row, several pedestrians are crossing the street, which are mostly segmented by our DNN, whereas the baseline DNN mostly misses the persons in the center as well as all heads. In the bottom row, the person in front of the car is completely overlooked by the baseline, and also some cars in the background are missed. 

When examining the OoD masks, we observed that the connected components are often very extensive, which is caused by neighboring OoD objects. Thus, the embedding space contains many large image patches which are not tailored to a single OoD object, but rather to a number of parked cars, a crowd of people or even a bicyclist riding next to a car, which appreciably impairs our results.

\section{Conclusion \& Outlook}

In our work, we proposed a solution to open world classification for image classification and semantic segmentation by learning novel classes in an unsupervised manner. We suggested to postulate empty classes, which allow one to capture newly observed classes in an incremental learning approach. This way, we allow our model to detect new classes in a flexible manner, potentially whitewashing mistakes of previous OoD detectors.

As our method employs several hyperparameters, e.g.\ , to specify the number of novel empty classes, we envision an automatic derivation of the optimal number of new classes as future work. In this regard, replacing the Elbow method in the eventual clustering by more suitable criteria appears desirable \cite{Schubert2022StopUT}. Moreover, we shall investigate approaches to improve the generalizability of our approach to embedding models of arbitrary kind to derive distance matrices, not being tailored to specific datasets. \SU{Furthermore, the semantic segmentation performance could be improved by incorporating depth information into the OoD segmentation method to obtain OoD candidates on instance- instead of segment-level.}

{\small
\bibliographystyle{ieee_fullname}
\bibliography{egbib}
}

\newpage
\section*{Appendix}
\appendix

\section{More Details on Experiments}
\subsection{TwoMoons}

As a proof of concept, consider a simple binary classification problem in the plane. As in-distribution data $1000$ samples are drawn from the Two Moons dataset\footnote{\label{sklearn}\href{https://scikit-learn.org/stable/modules/classes.html\#module-sklearn.datasets}{https://scikit-learn.org/stable/modules/classes.html\#module-sklearn.datasets}} with noise $= 0.1$. Additionally, $100$ OoD data samples are drawn from a uniform distribution over $[-4,4]^2$, as illustrated in \cref{fig:toy_example}. Then, a shallow neural network, consisting of $4$ fully connected layers, is trained on these samples to minimize the cross entropy with respect to the Two Moons data while maximizing the entropy on the OoD data. As test data, another $750$ samples are drawn from the Two Moons dataset, together with $500$ OoD samples belonging to three blobs, centered at $c_1=(-1.5,-0.95), c_2=(2.5,1.5)$ and $c_3=(3,-1)$, respectively, with $0.25$ standard deviation. These blobs represent the novel classes. The test data is then fed into the trained model, and is considered to be OoD if the softmax entropy exceeds a threshold of $0.8$. Finally, the initial model is extended by three empty classes in the last layer and then fine-tuned on the Two Moons training samples plus the OoD samples detected in the test data. The OoD data is clustered into the empty classes through our proposed loss function, without requiring any previous (pseudo-)labeling. 

\subsection{Training Parameters}

We provide an overview of the training parameters for each experiment in~\cref{tab:params}. We performed experiments using the Adam and the SGD optimizer as well as different batch sizes. Only the batch size for semantic segmentation was bounded by memory limitations. The hyperparameters which are related to the loss functions, namely $\alpha,\lambda_1,\lambda_2,\lambda_3$, were selected by trying out and monitoring the course of the loss functions.

\section{Evaluation Metrics for Randomized OoD Classes}

Our approach assumes, that novel classes are well separable in the feature space. Thus, we have selected suitable classes by hand. In~\cref{tab:results_ic_mean}, we provide evaluation metrics which are averaged over 5 runs with randomly selected classes, respectively. In particular, for FashionMNIST, we observe a large standard deviation of $27.61\%$ in the accuracy of novel classes. Our method fails, whenever the novel classes are \emph{t-shirt/top} $(0)$ or \emph{shirt} $(6)$, which are semantically similar and also not separable in the feature space. The same holds for Animals10 regarding the classes \emph{horse}, $(1)$, \emph{cow} $(6)$ and \emph{sheep} $(7)$.

\section{OoD Detection}

\begin{figure}[t]
    \centering
    \captionsetup[subfloat]{labelformat=empty}
    \subfloat[MNIST: $0,5,7$]{\includegraphics[width=0.23\textwidth]{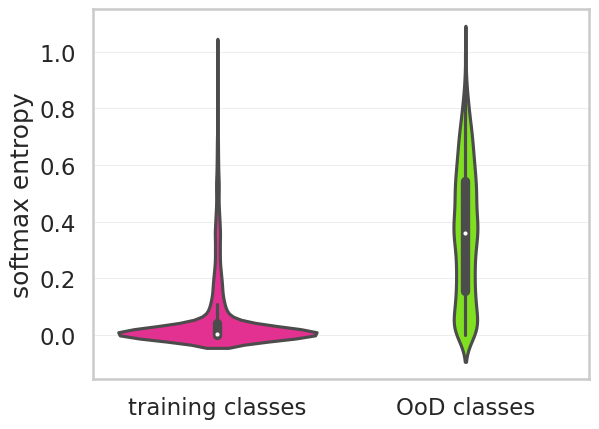}}\hfill
    \subfloat[FashionMNIST: $1,8$]{\includegraphics[width=0.23\textwidth]{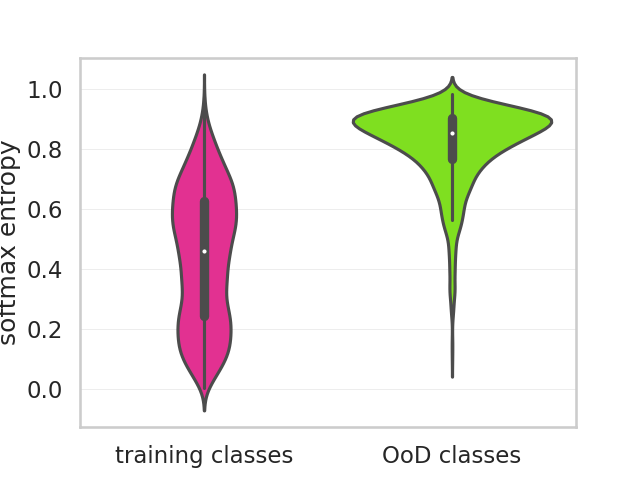}}\\
    \subfloat[Cifar10: $11,12$]{\includegraphics[width=0.23\textwidth]{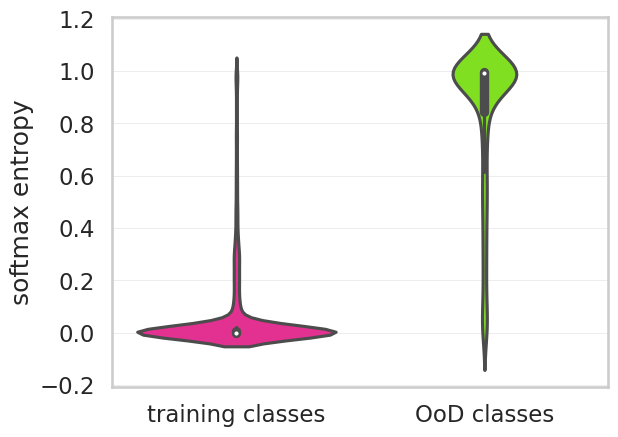}}\hfill
    \subfloat[Animals10: $3,4,8,9$]{\includegraphics[width=0.23\textwidth]{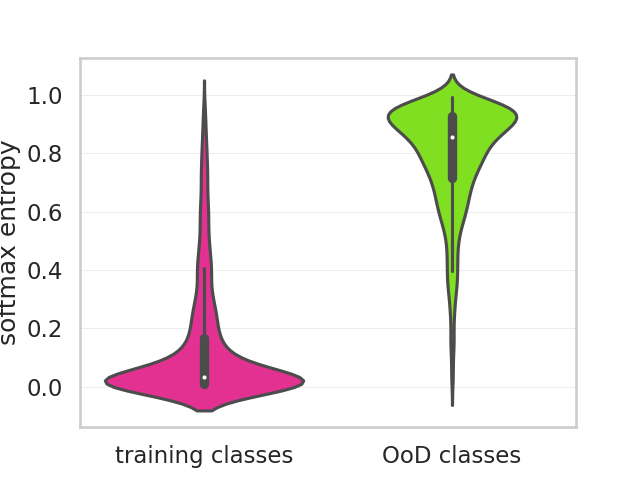}}
    \caption{Visualization of the softmax entropy, that the initial models exhibit on samples of known and OoD classes, respectively.}
    \label{fig:violines}
\end{figure}

\begin{figure}[t]
    \centering
    \captionsetup[subfloat]{labelformat=empty}
    \subfloat[MNIST: $0,5,7$]{\includegraphics[width=0.23\textwidth]{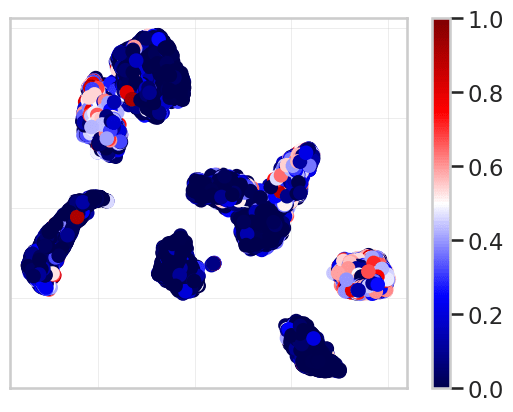}}\hfill
    \subfloat[FashionMNIST: $1,8$]{\includegraphics[width=0.23\textwidth]{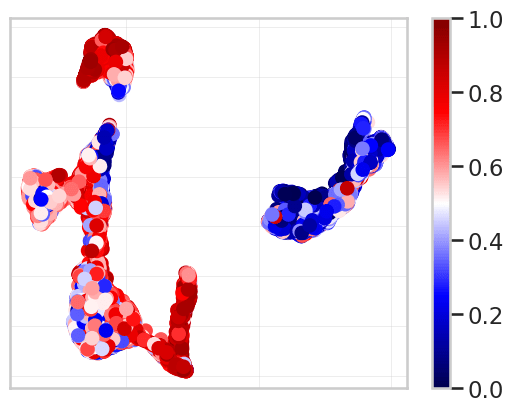}}\\
    \subfloat[Cifar10: $11,12$]{\includegraphics[width=0.23\textwidth]{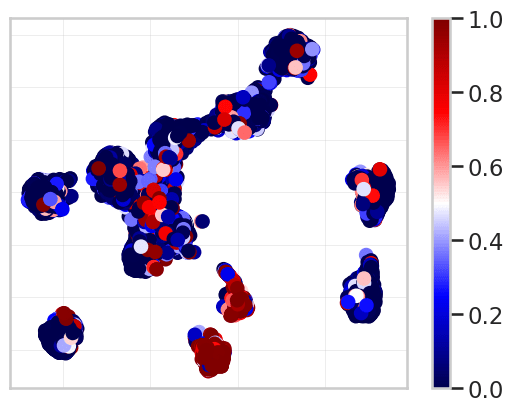}}\hfill
    \subfloat[Animals10: $3,4,8,9$]{\includegraphics[width=0.23\textwidth]{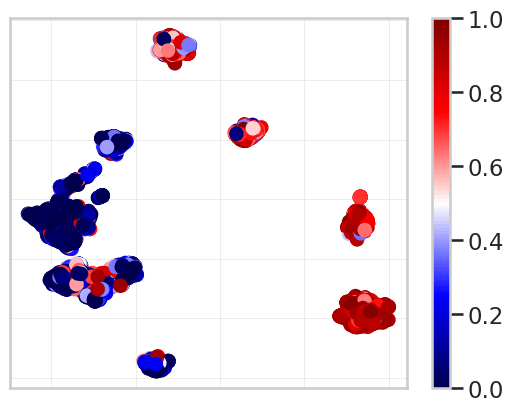}}
    \caption{Visualization of the softmax entropy per data sample, that the initial models exhibits on test samples.}
    \label{fig:entropy}
\end{figure}

\begin{figure*}[t]
    \captionsetup[subfloat]{labelformat=empty}
    \centering
    \subfloat[]{ \includegraphics[width=0.24\textwidth]{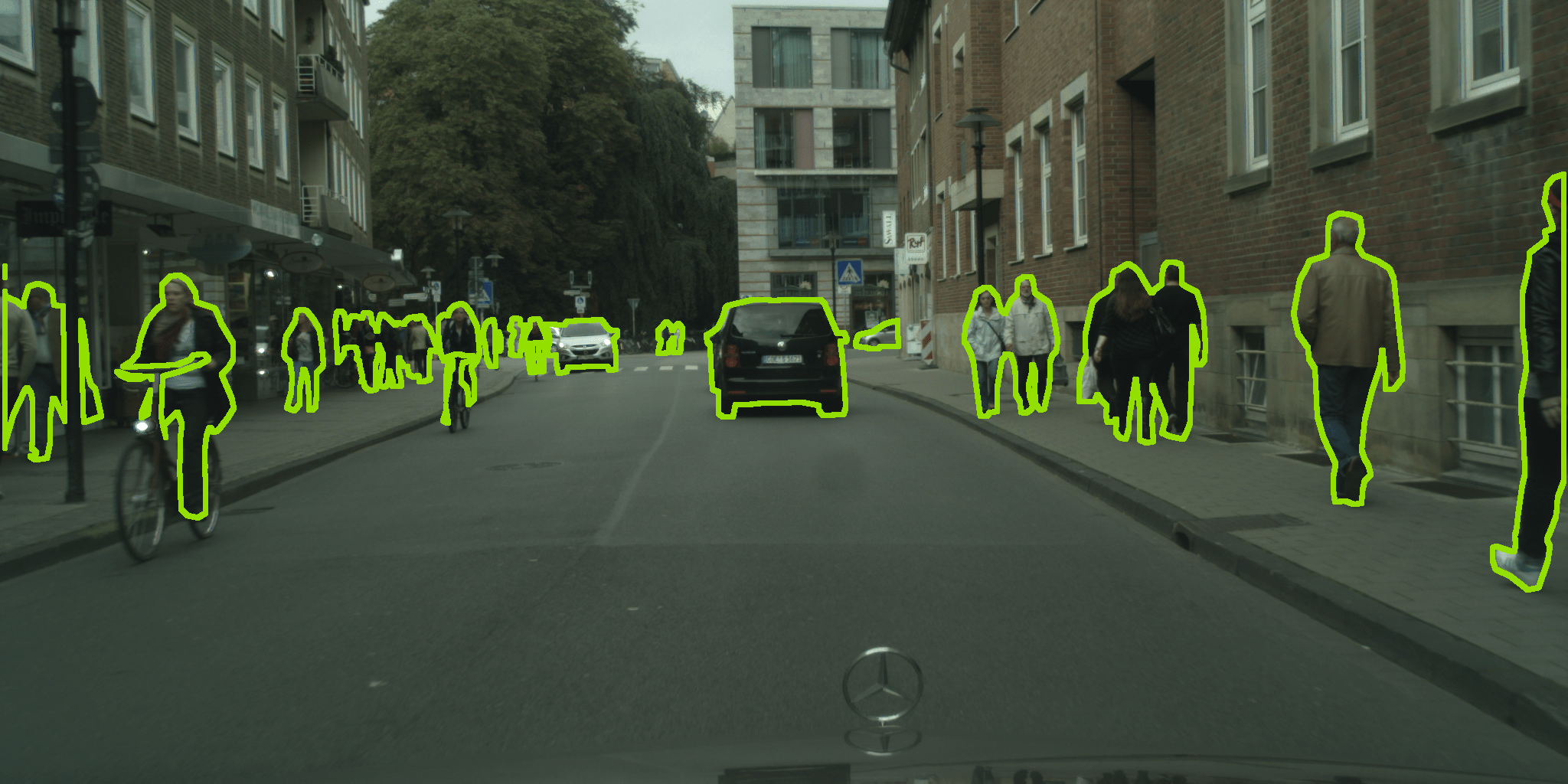}}\hfill
    \subfloat[]{ \includegraphics[width=0.24\textwidth]{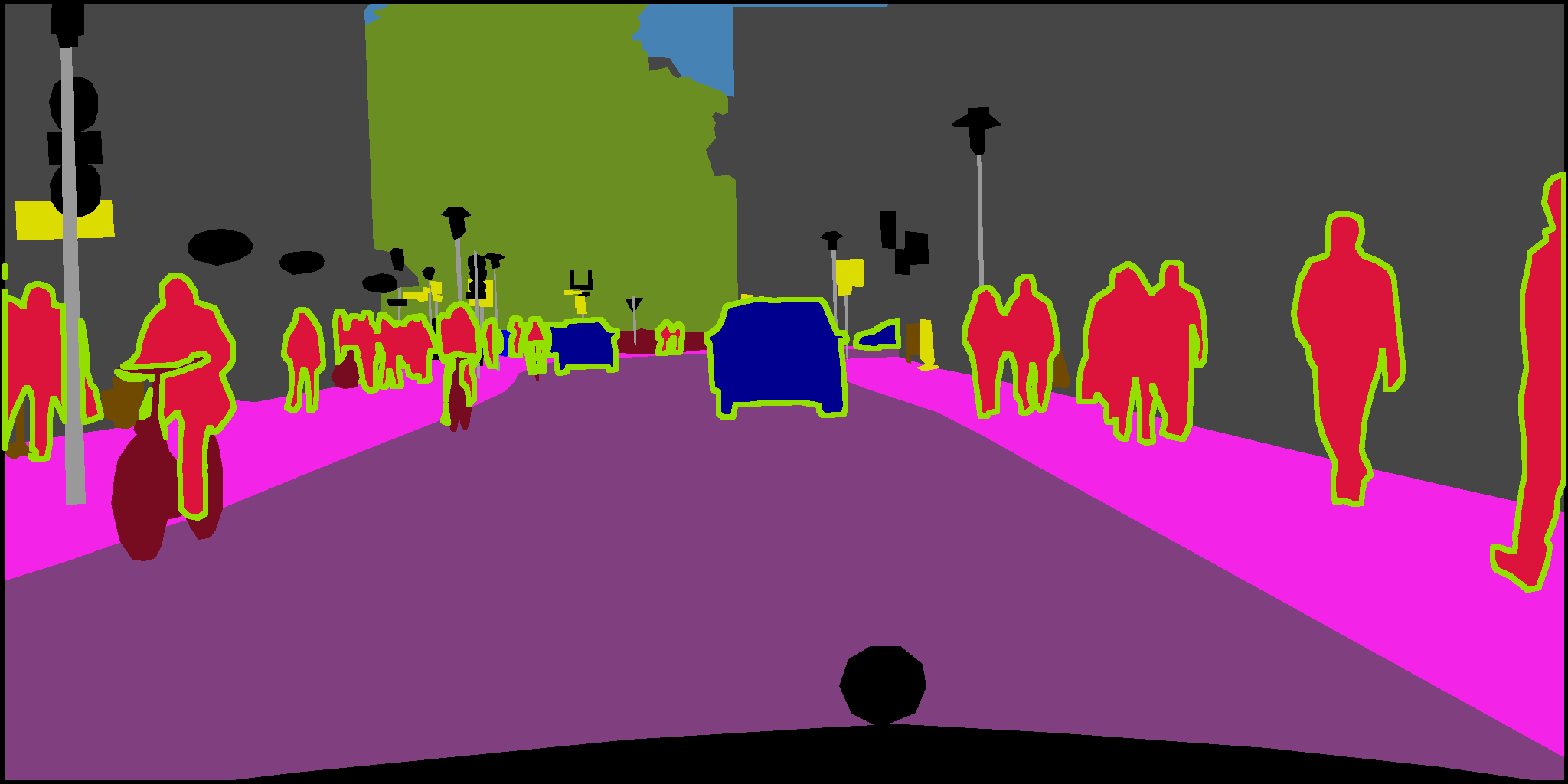}}\hfill
    \subfloat[]{ \includegraphics[width=0.24\textwidth]{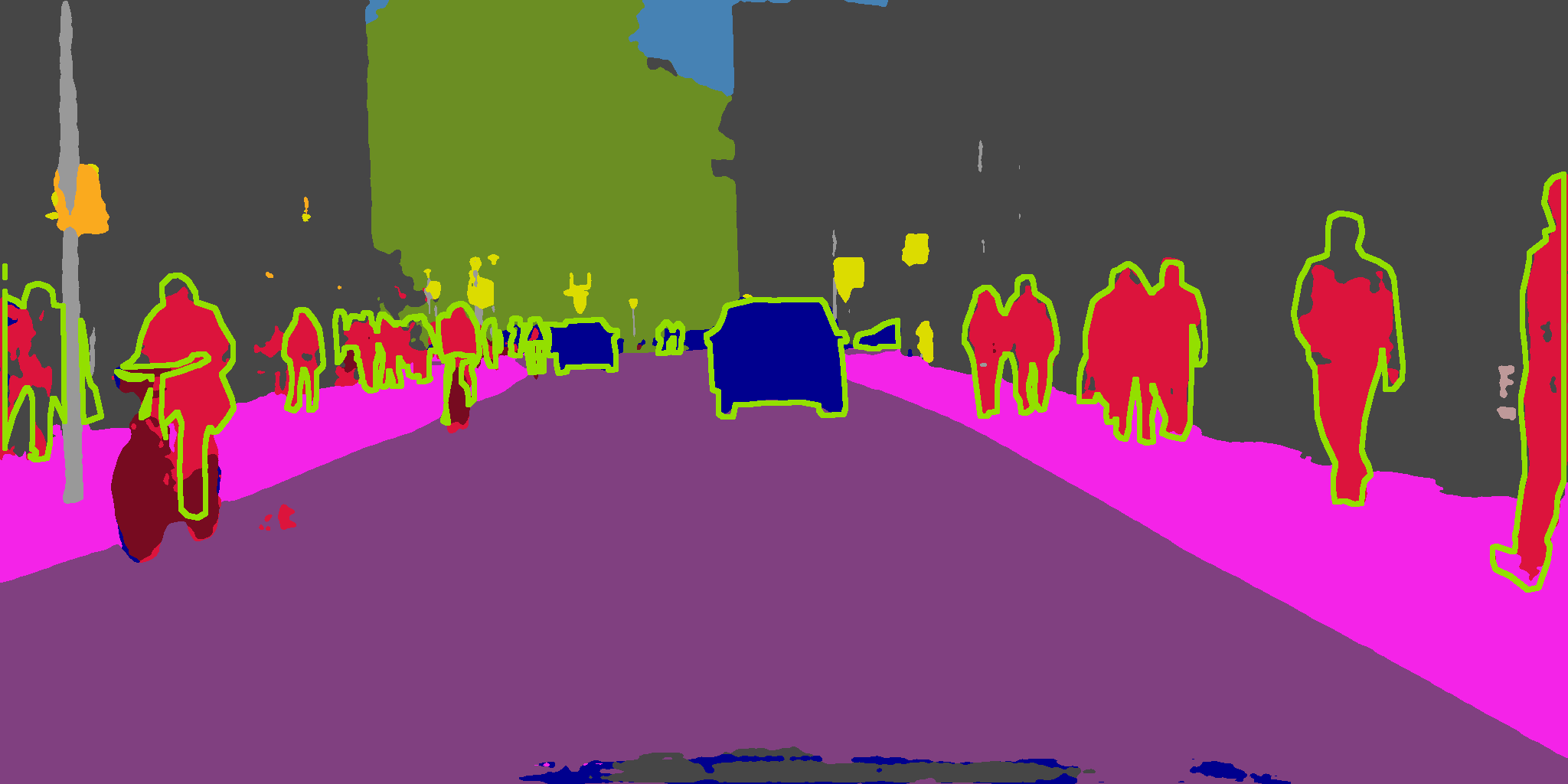}}\hfill
    \subfloat[]{ \includegraphics[width=0.24\textwidth]{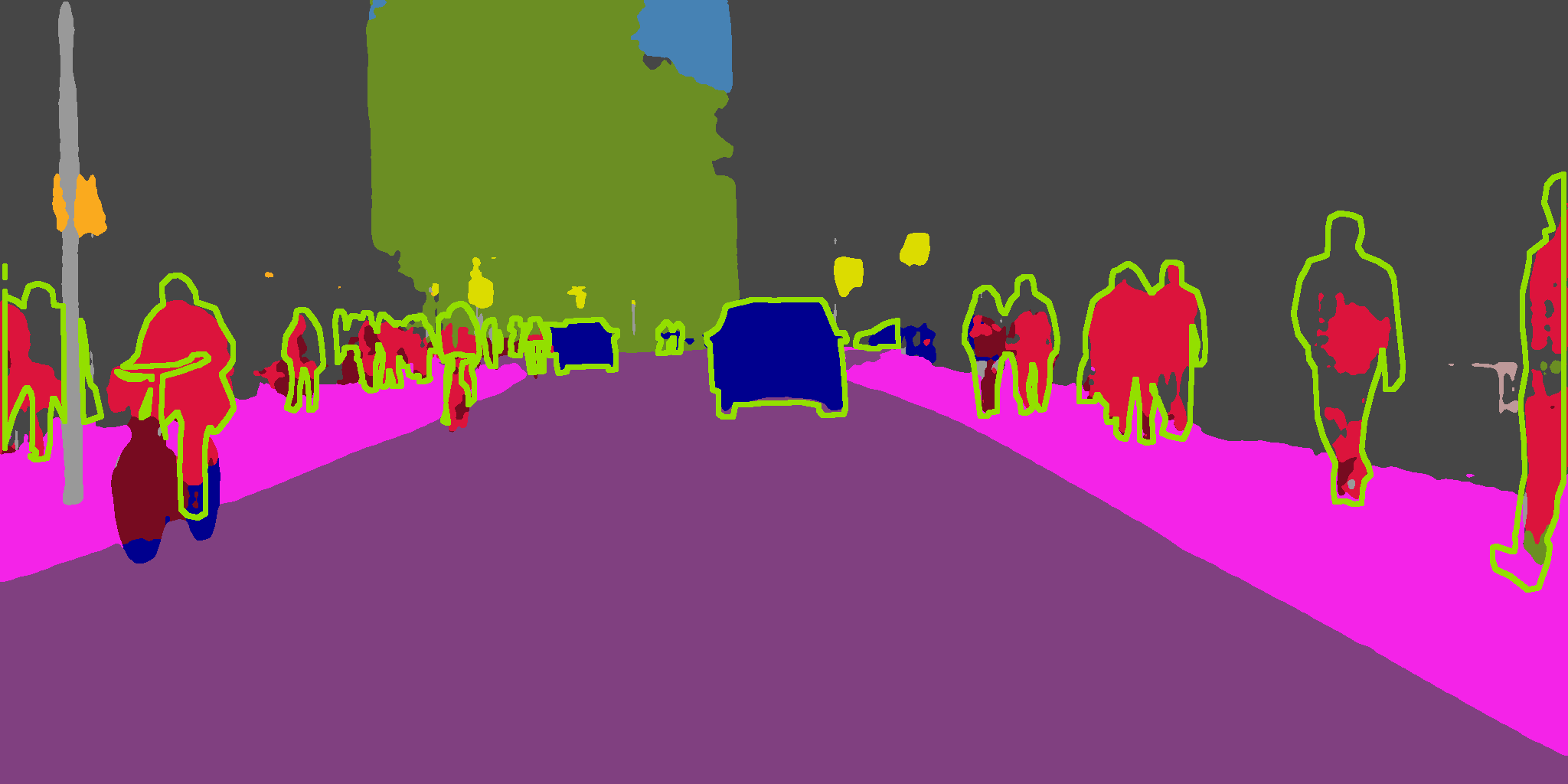}}\\[-2em]
    \subfloat[image]{ \includegraphics[width=0.24\textwidth]{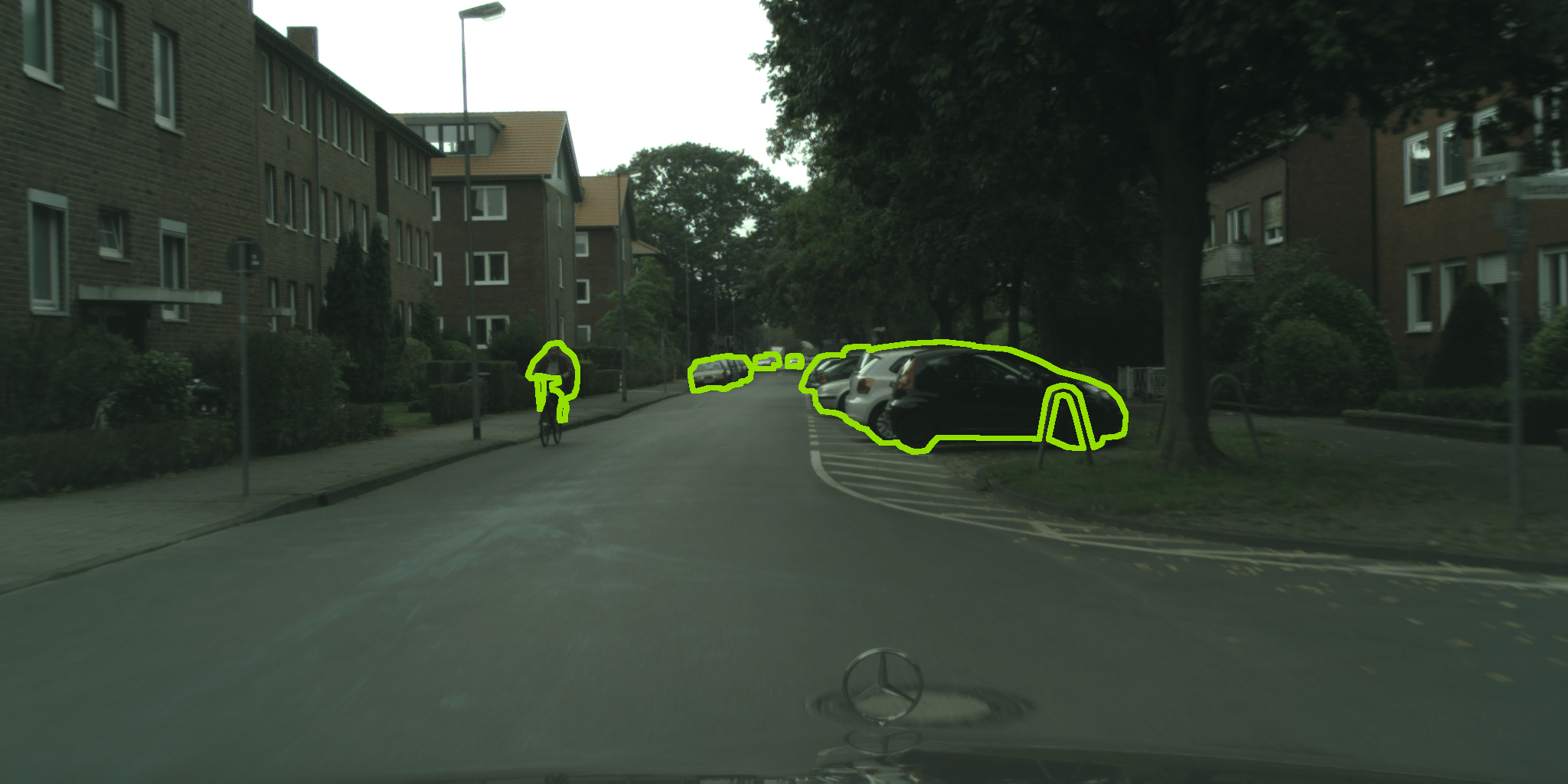}}\hfill
    \subfloat[ground truth]{ \includegraphics[width=0.24\textwidth]{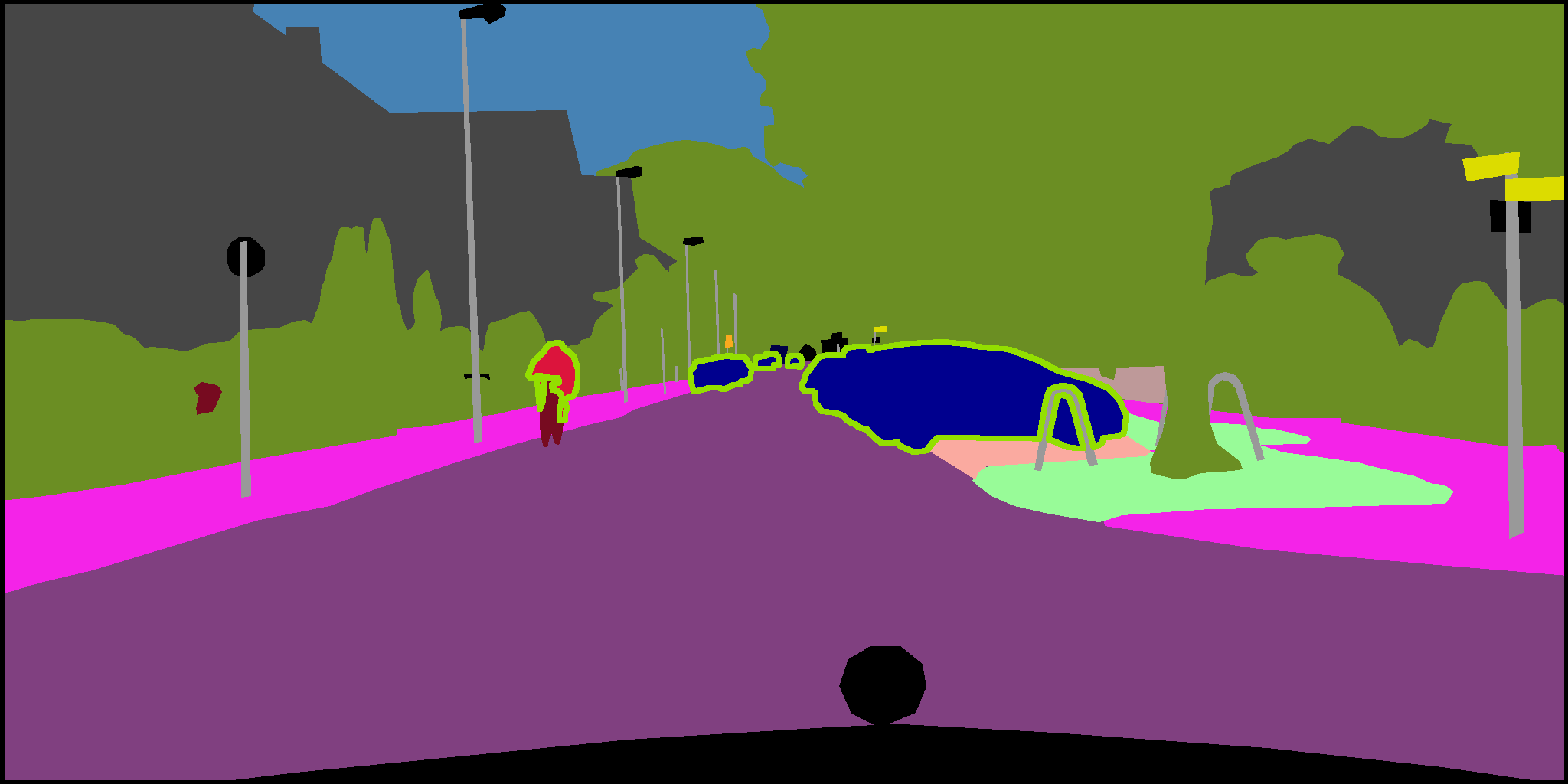}}\hfill
    \subfloat[ours]{ \includegraphics[width=0.24\textwidth]{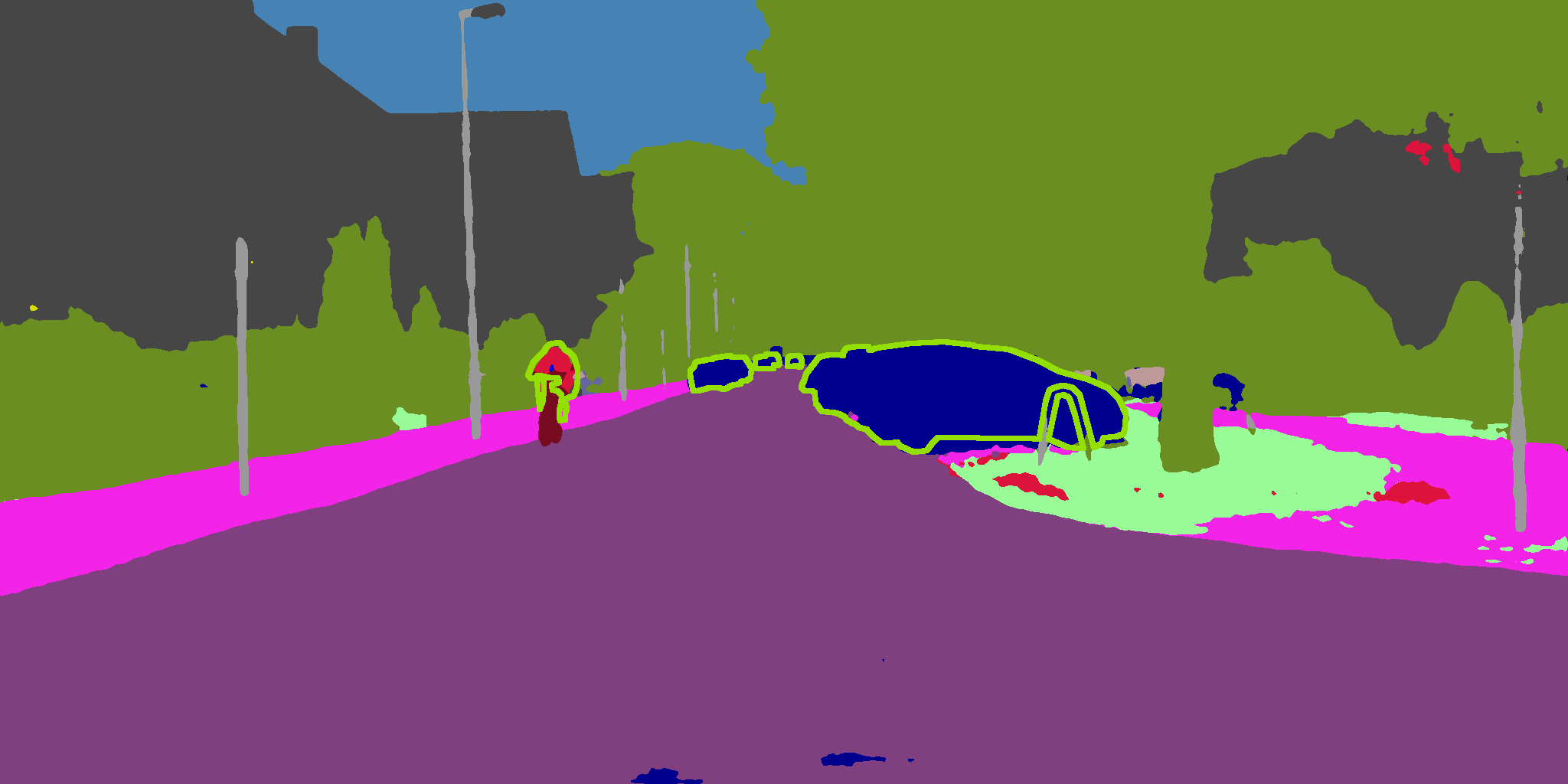}}\hfill
    \subfloat[baseline]{ \includegraphics[width=0.24\textwidth]{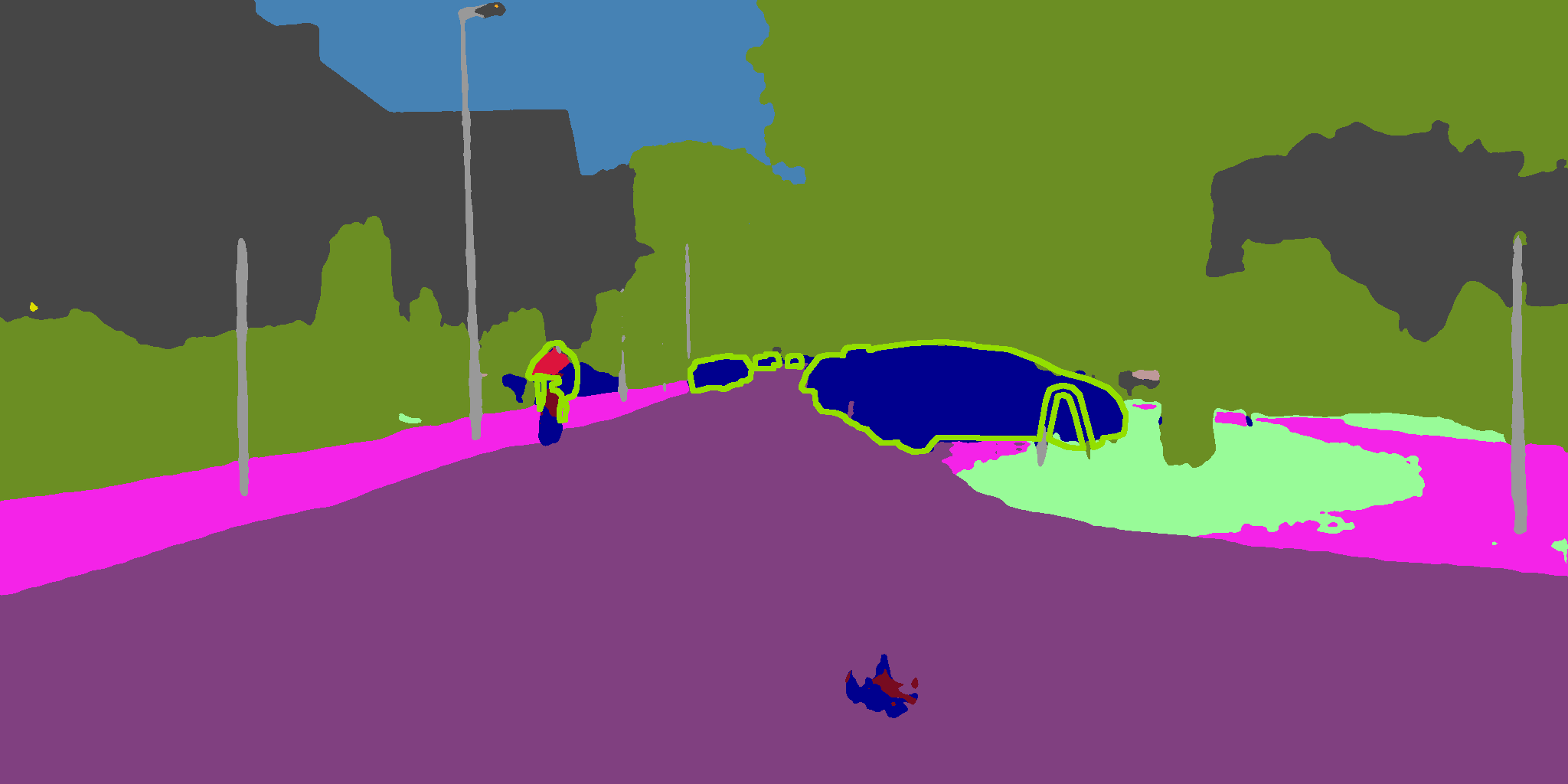}}
    \caption{Visual comparison of the segmentation masks produced by our method and by the baseline for two images from the Cityscapes validation dataset. The ground truth contours of the novel classes are highlighted with green.}\label{fig:more_results_cityscapes}
\end{figure*}

\begin{table*}[t]
    \centering
    \resizebox{1\textwidth}{!}{
    \begin{tabular}{l||c|c|c|c|c|c|c|c|c|c|c}
        dataset & $\#$ empty classes & $\#$ epochs & optimizer & learning rate & momentum & weight decay & batch size & $\alpha$ & $\lambda_1$ & $\lambda_2$ & $\lambda_3$ \\\hline
        MNIST & 3 & 30 & adam & 1e-2 & - & - & 2500 & 5 & 0.45 & 0.45 & 0.1 \\
        FashionMNIST & 2 & 30 & sgd & 1e-2 & 0 & 0 & 500 & 2.5 & 0.45 & 0.45 & 0.1 \\
        Cifar10 & 2 & 30 & sgd & 1e-2 & 0.9 & 1e-4 & 1000 & 5 & 0.45 & 0.45 & 0.1 \\
        Animals10 & 4 & 30 & adam & 5e-3 & - & - & 1000 & 2.5 & 0.45 & 0.45 & 0.1 \\
        Cityscapes & 2 & 200 & adam & 5e-3 & - & - & 10 & 2.5 & 0.375 & 0.375 & 0.25 \\
    \end{tabular}}
    \caption{Overview of training parameters for each dataset.}
    \label{tab:params}
\end{table*}

For image classification, we employed entropy maximization during training the initial model. The softmax entropy for the test data is visualized as a summary statistic in~\cref{fig:violines} and sample-wise in~\cref{fig:entropy}. We observe that the DNN exhibits high entropy scores on OoD data for all datasets except MNIST. However, the entropy for MNIST in-distribution data is sufficiently small, so that we detect most OoD samples using a threshold of $\tau=0.1$. Further, the initial FashionMNIST DNN is uncertain regarding the in-distribution classes \emph{t-shirt/top} $(0)$, \emph{pullover} $(2)$ and \emph{shirt} $(6)$. However, this may be aleatoric uncertainty. To avoid too many false positive OoD predictions, we choose a high threshold $\tau=0.75$. For the remaining datasets, in-distribution and OoD samples are well separable by the softmax entropy, thus, there is a large interval of proper thresholds.

\vspace{10cm}

\begin{table}[h]
    \centering
    \resizebox{1\linewidth}{!}{
    \begin{tabular}{l|c|l||c|c}
        \hline
        \multicolumn{5}{c}{\textbf{Image Classification}}\\\hline
        \textbf{dataset} & $\#$\textbf{OoD} & \textbf{accuracy} & \textbf{initial} &  \textbf{ours} \\\hline\hline
        % \multirow{2}{*}{MNIST} & \multirow{2}{*}{3} & known & $\%$  & \cellcolor{gray!25}$\%$ \\
        % & & novel & - & \cellcolor{gray!25}$\%$ \\\hline
        \multirow{2}{*}{FashionMNIST} & \multirow{2}{*}{2} & known & $84.72 \pm 03.25\%$  & \cellcolor{gray!25}$82.78 \pm 02.18\%$\\
        & & novel & - & \cellcolor{gray!25}$63.42 \pm 27.61\%$ \\\hline
        % \multirow{2}{*}{CIFAR10} & \multirow{2}{*}{} & known & $\%$ & \cellcolor{gray!25}$\%$\\
        % & & novel & - & \cellcolor{gray!25}$\%$\\\hline
        \multirow{2}{*}{Animals10} & \multirow{2}{*}{4} & known & $97.02 \pm 00.63\%$ & \cellcolor{gray!25}$94.70 \pm 00.59\%$\\
        & & novel & - & \cellcolor{gray!25}$66.79 \pm 25.82\%$\\\hline
    \end{tabular}}
    \caption{Quantitative evaluation of the FashionMNIST and Animals10 experiments, averaged over $5$ runs with randomly selected OoD classes, each. For all evaluated models, the accuracy is stated separately for the previously-known and the unlabeled novel classes.}
    \label{tab:results_ic_mean}
\end{table}

\section{Semantic Segmentation}

\begin{figure}[t]
    \captionsetup[subfloat]{labelformat=empty}
    \centering
    \subfloat[predicted segmentation]{ \includegraphics[width=0.23\textwidth]{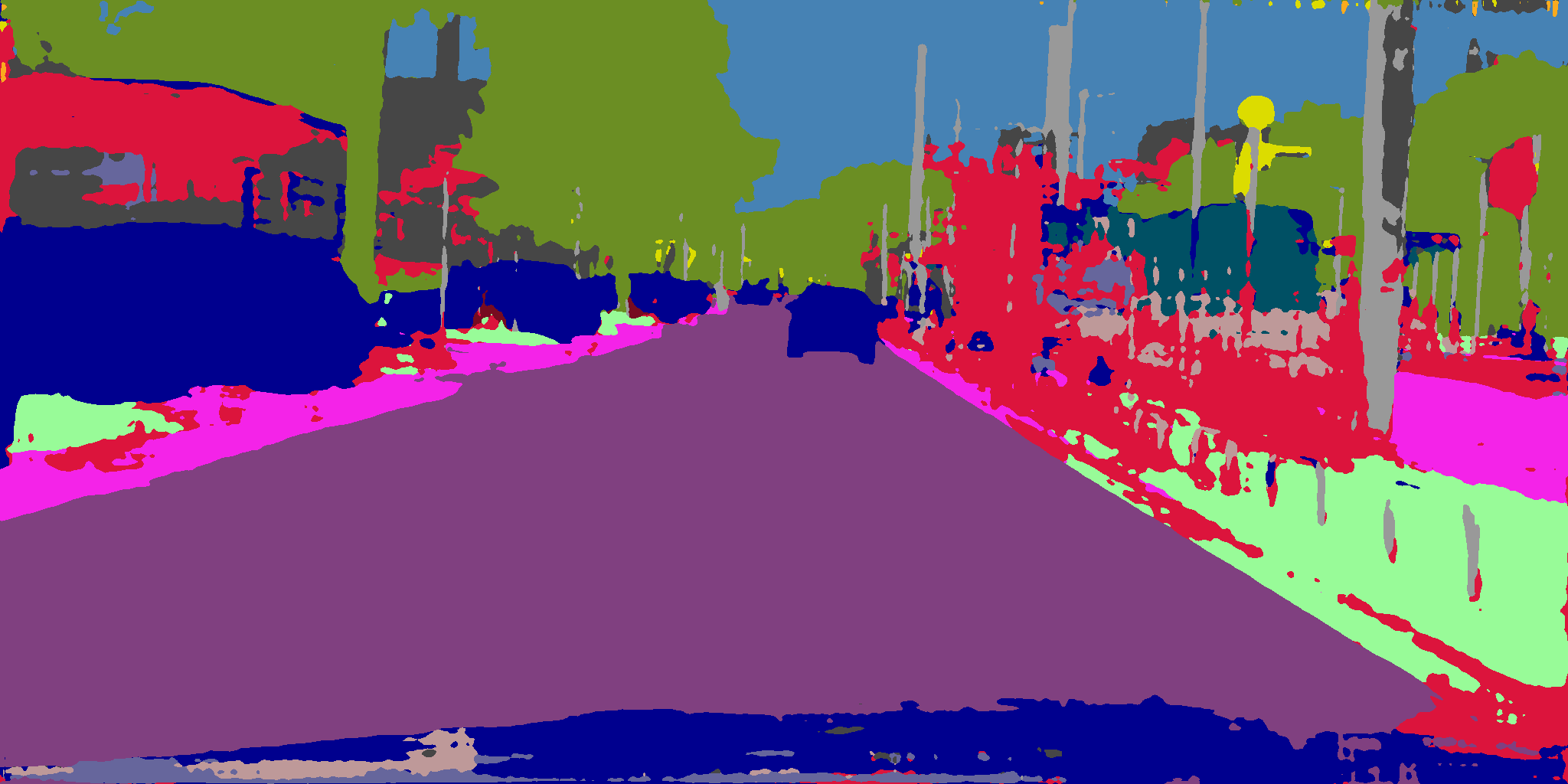}}\hfill
    \subfloat[softmax entropy]{ \includegraphics[width=0.23\textwidth]{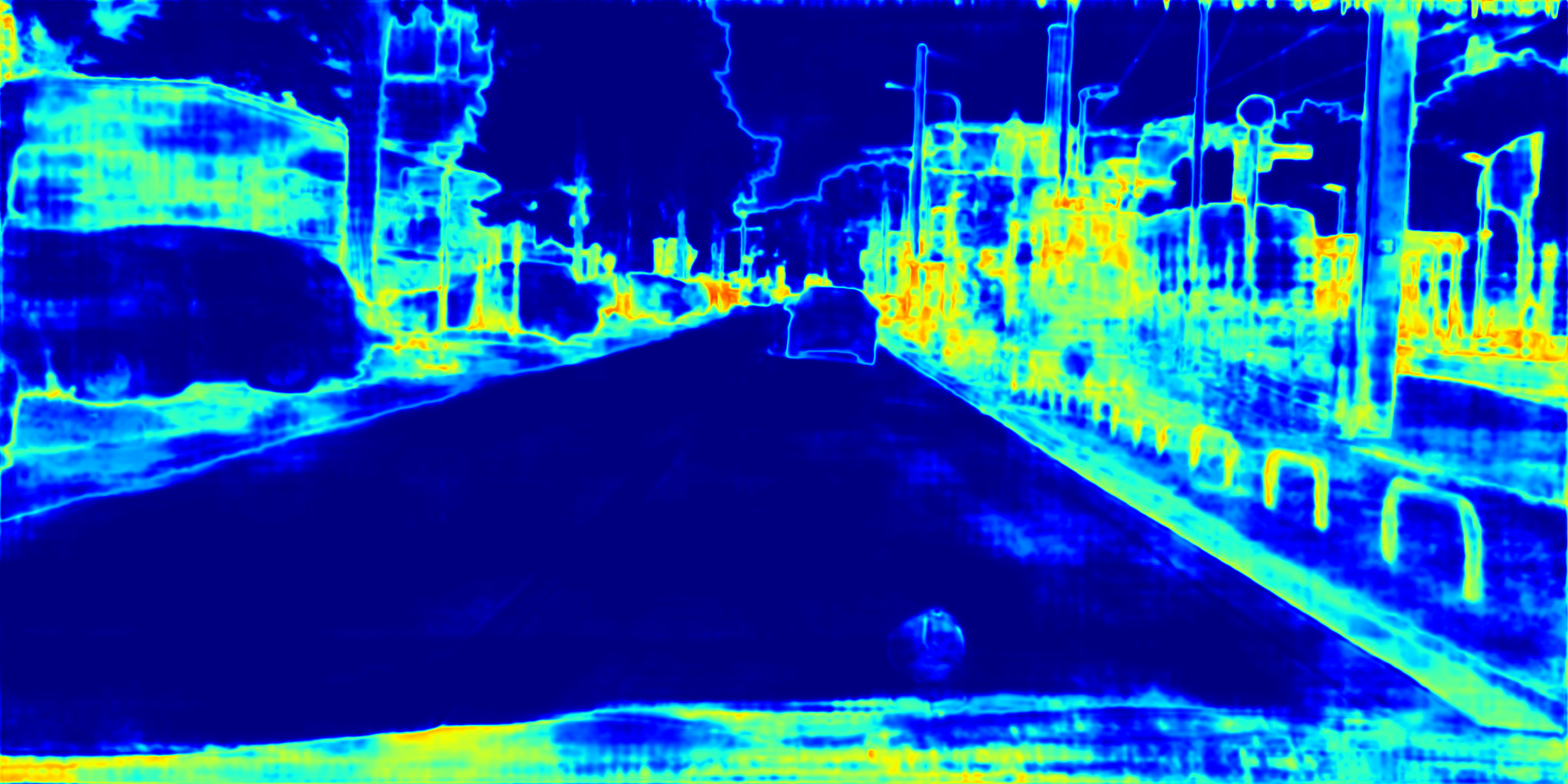}}
    \caption{For highly uncertain regions, the extended DNN tends to predict the novel \emph{human} class, which causes the low precision score.}\label{fig:fp_cityscapes}
\end{figure}

We provide some more qualitative semantic segmentation results in~\cref{fig:more_results_cityscapes}, for which our method outperforms the baseline. Furthermore, we also give an example in~\cref{fig:fp_cityscapes}, where regions which come along with a high softmax entropy are wrongly predicted as the novel class \emph{human}.

\end{document}